\documentclass[runningheads]{llncs}
\usepackage{graphicx}

\usepackage{tabularx}
\usepackage{multirow}
\usepackage{booktabs}
\usepackage{kotex}
\usepackage{wrapfig}
\usepackage[font=footnotesize]{caption}
\usepackage{algorithm}
\usepackage[noend]{algpseudocode}
\usepackage[final]{pdfpages}

\usepackage{tikz}
\usepackage{comment}
\usepackage{amsmath,amssymb} 
\usepackage{color}

\usepackage[accsupp]{axessibility}  

\begin{document}
\pagestyle{headings}
\mainmatter
\def\ECCVSubNumber{3216}  

\title{Controllable Image Enhancement} 

\titlerunning{Controllable Image Enhancement}
%
\author{Heewon Kim \and
Kyoung Mu Lee }
\authorrunning{H. Kim et al.}
%

\institute{ASRI, Department of ECE, Seoul National University\\
\email{\{ghimhw, kyoungmu\}@snu.ac.kr}}
\maketitle

\begin{abstract}
   Editing flat-looking images into stunning photographs requires skill and time.
   Automated image enhancement algorithms have attracted increased interest by generating high-quality images without user interaction.
   However, the quality assessment of a photograph is subjective.
   Even in tone and color adjustments, a single photograph of auto-enhancement is challenging to fit user preferences which are subtle and even changeable.
   To address this problem, we present a semi-automatic image enhancement algorithm that can generate high-quality images with multiple styles by controlling a few parameters.
   We first disentangle photo retouching skills from high-quality images and build an efficient enhancement system for each skill.
   Specifically, an encoder-decoder framework encodes the retouching skills into latent codes and decodes them into the parameters of image signal processing (ISP) functions.
   The ISP functions are computationally efficient and consist of only 19 parameters.
    Despite our approach requiring multiple inferences to obtain the desired result,
    experimental results present that the proposed method achieves state-of-the-art performances on the benchmark dataset for image quality and model efficiency.

\end{abstract}

\section{Introduction}
\label{sec:introduction}
As most people have their own digital camera, they enjoy taking pictures of precious moments.
However, obtaining high-quality photographs is challenging.
While tone and color adjustment benefit visual attractiveness, manually retouching photographs requires expertise and effort using professional tools (e.g., Photoshop).
More importantly, the looking better photograph is subjective and difficult to define.
Photographers retouch an image in different ways~\cite{fivek}, while ordinary people have rather ambiguous preferences.

Most works on auto-enhancement assume that people have distinct preferences~\cite{kang2010personalization,caicedo2011collaborative} and aim to learn the characteristics of high-quality photographs for specific preferences.
For instance, recent approaches~\cite{chen2018dpe,wang2019upe,moran2020depplpf,wang2021realtime,jie2021LPTN} employ deep neural networks to transform a flat-looking image to the photograph retouched by an expert~\cite{fivek}.
The works on personalizing enhancement algorithms~\cite{kang2010personalization,caicedo2011collaborative,kim2020pienet} allow users to select preferred retouching styles and transform all test images to the selected styles.
However, the user preferences are subtle.
Users are ambivalent to different styles of a photograph and often change their minds unexpectedly. 
Learning such an ambiguous goal of ``preferences'' may limit the potential of deep learning approaches.

\begin{figure}[t]
	\centering
	\includegraphics[width=1\linewidth]{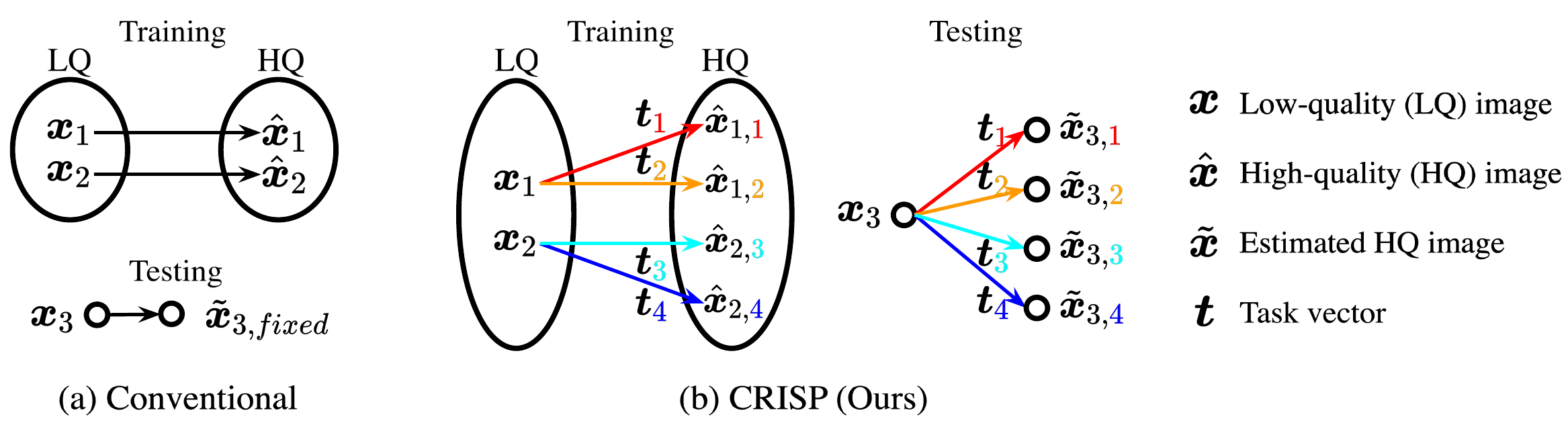} \\ \vspace{-0.3cm}
	\scriptsize
	\caption{Overview of image enhancement algorithms. 
	(a) Conventional learning-based approaches (\textit{e.g.}, DPE~\cite{chen2018dpe}) regard that training data has a single style of high-quality (HQ) images ($\hat{\boldsymbol{x}}_{1}$ and $\hat{\boldsymbol{x}}_{2}$) and transform a test image ($\boldsymbol{x}_{3}$) to the style ($\tilde{\boldsymbol{x}}_{3,fixed}$).
	(b) We assume that a low-quality (LQ) image ($\boldsymbol{x}_{1}$) has multiple HQ images with different styles ($\hat{\boldsymbol{x}}_{1,{\color{red}1}}$ and $\hat{\boldsymbol{x}}_{1,{\color{orange}2}}$). Our proposed scheme (CRISP) learns the mappings for distinct styles represented by task vector ($\boldsymbol{t}$). In testing, CRISP can generate multiple styles of HQ images ($\tilde{\boldsymbol{x}}_{3,{\color{red}1}}$, $\tilde{\boldsymbol{x}}_{3,{\color{orange}2}}$, $\tilde{\boldsymbol{x}}_{3,{\color{cyan}3}}$, and $\tilde{\boldsymbol{x}}_{3,{\color{blue}4}}$) by adjusting task vector.
	}
	\vspace{-1em}
	\label{fig:overview}
\end{figure}

Instead, this paper covers a controllable image enhancement scenario in which the user selects the desired result in various styles for a test image.
We seek to disentangle retouching skills from high-quality photographs and build a semi-automatic image enhancement system that generates diverse styles for a flat-looking image.
The key points that we wish to address are: 
(1) Is it possible to identify, imitate, and generalize retouching skills for high-quality photographs? 
(2) How can non-experts find the desired results in a feasible time?

To this end, first, we regard an image enhancement process as an one-to-many mapping; an image has multiple possible solutions (or retouching styles) of high-quality photographs.
Existing methods~\cite{chen2018dpe,wang2019upe,moran2020depplpf,wang2021realtime,jie2021LPTN} trained their models from a flat-looking photograph to a visually pleasing one as one-to-one mapping, which led the models to generate an average (or mode collapse) output of the possible solutions (See Figure~\ref{fig:overview}{\color{red}(a)}).
In contrast, we believe that each solution has an identical retouching skill.
Using both an image and a retouching skill (represented by a task vector) as inputs, we can reformulate the image enhancement process as one-to-one mapping, as illustrated in Figure~\ref{fig:overview}{\color{red}(b)}.

Second, we conduct a lightweight and computationally efficient enhancement system using the image signal processing~(ISP) pipeline of digital camera~\cite{ramanath2005color}.
We reinterpret the ISP pipeline as an image retouching process, where ISP conventionally renders human-readable RGB images from raw sensor data.
ISP has been widely used in real-world digital cameras due to its few parameters and negligible computation of simple color transformations (\textit{e.g.}, scaling by a scalar value).
While camera manufacturers carefully tune the ISP parameters based on the principles of an image sensor and human perception of colors, we tune the ISP parameters for image enhancement.

Specifically, we present ContRollable Image Signal Processing, referred to as CRISP, that reparameterizes a general ISP pipeline by an encoder-decoder framework.
During training, a convolutional neural network encodes the retouching skills into a control factor (or task vector), which users adjust in test time.
Decoder consists of fully connected layers and predicts the parameters of the ISP pipeline from the control factor.

Experimental results present that CRISP can generate \textit{crisp} photographs with diverse retouching styles.
CRISP outperforms the state-of-the-art methods in MOS, PSNR, and SSIM on MIT-Adobe FiveK benchmark datasets.
CRISP uses 2$\times$ fewer parameters and 100$\times$ smaller FLOPs per inference compared to the existing methods due to its 19 parameters of the ISP pipeline. 
Moreover, CRISP can reduce the dimension of the control factor (or task vector) to 3, small enough to be adjusted at test time.
A simple greedy algorithm finds the task vector that performs the state-of-the-art image quality in five steps.

\section{Related Works}
\label{sec:related_works}
Image enhancement has a long and rich history of research.
While professional image editing tools (e.g., Photoshop) offer a wide range of control and flexibility to adjust tone and color, we categorize image enhancement algorithms according to the degree of automation.
The first category includes most enhancement approaches aimed at universal automation.
These approaches include early works that enhance image contrast using histogram equalization~\cite{pizer1987adaptive}, gamma correction~\cite{huang2013efficient}, or retinex theory~\cite{land1971lightness}.
Data-driven approaches~\cite{hwang2012context,fivek,yan2014rank} learn contrast, color, and brightness adjustments from a large dataset used for general high-quality photographs.
Deep-learning-based methods~\cite{chen2018dpe,moran2020depplpf,wang2019upe,Kim2021representative,kim2020global} have achieved breakthroughs in this area.
Recent works~\cite{jie2021LPTN,wang2021realtime,he2020conditional} focused on improving the efficiency of neural networks due to their high computation costs.
However, the automatically enhanced images have limitations in satisfying users with different preferences.

Second, the line of works~\cite{kang2010personalization,joshi2010personal,hsu2008light,caicedo2011collaborative,kim2020pienet} aim to personalize an auto-enhancement algorithm.
They allow users to select preferred images first and retouch all test images similar to the styles with the selected ones.
However, these approaches share the limitation of universal automation, as the preferred retouching style depends on image content and can be changed unexpectedly.

\begin{figure*}[t]
	\centering
	\includegraphics[width=1\linewidth]{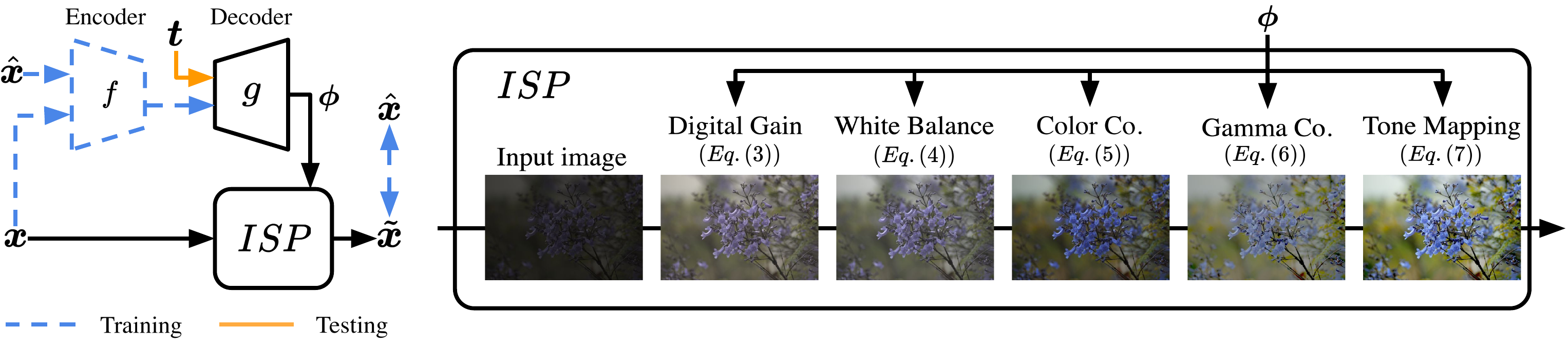} \\ \vspace{-0.1cm}
	\scriptsize
	\caption{Overview of the proposed controllable image signal processing (CRISP) pipeline. (left) CRISP generates the \textit{task-adaptive} ISP parameters ($\boldsymbol{\phi}$) using an encoder-decoder framework. In training, the encoder disentangles the retouching skills from high-quality images ($\hat{\boldsymbol{x}}$), while users adjust the retouching skills by controlling the values of the task vector ($\boldsymbol{t}$) for a test image. (right) Our image signal processing (ISP) consists of simple arithmetic functions with 19 parameters to estimate HQ images.
	}
	\vspace{-1em}
	\label{fig:framework}
\end{figure*}

The third and last category allows users to interactively adjust each image input.
Works in this area aim to reduce the time and effort of using professional image editing tools but have attracted less attention despite its importance.
Lischinski et al.~\cite{lischinski2009interactive} presented a local tone adjustment for the regions selected by users.
Xiao et al.~\cite{xiao2018brightness} proposed histogram-based algorithms to control brightness and contrast for image enhancement.
Zero-DCE~\cite{Zero-DCE} estimates light-enhancement curves that allow user interaction by adjusting parameters.
However, these approaches did not address the end-to-end process of tone and color adjustment.
The work closest to ours is CSRNet~\cite{he2020conditional} which demonstrated image interpolation using affine combination of two different retouching styles.
By contrast, the proposed algorithm can explore a variety of professional retouching styles by controlling a few parameters.

\section{Controllable Image Enhancement Scenario}
\label{sec:CRISP_scenario}
Diverse user preferences over images have recently led to a great amount of interest in \textit{controllable} image manipulation~\cite{jingwen2019modulating,jingwen2020interactive,kim2021searching,xintao2019deep,he2020conditional}, aiming to adjust the outputs of algorithms by controlling some factors.
Our approach is an application of this area that adjusts photographs for enhancement by controlling the parameters of an image signal processing (ISP) pipeline.
Direct parameter tuning to the ISP pipeline may require skill and time similar to an image retouching process of photographers.
For the easy and fast adaptation, we reduce the dimension of the control factor by an encoder-decoder framework and reparameterize the ISP pipeline by the latent code of the framework, as described in Figure~\ref{fig:framework}.

Formally, the encoder $f$ represents the retouching skills of photographs in the $D$-dimensional latent code using the low-quality image $\boldsymbol{x}$ and the paired retouched high-quality image $\hat{\boldsymbol{x}}$ as inputs. 
The decoder $g$ predicts the \textit{task-adaptive} parameters of an ISP pipeline $\boldsymbol{\phi}$.
The ISP pipeline ($ISP$) generates retouched photographs ($\tilde{\boldsymbol{x}}$) from the low-quality image as follows,
\begin{equation}\label{eq:isp}
\tilde{\boldsymbol{x}} = 
ISP(\boldsymbol{x};\boldsymbol{\phi}) =
\begin{cases}
ISP(\boldsymbol{x};g(f(\boldsymbol{x},\hat{\boldsymbol{x}})) & \mbox {if training} \\
ISP(\boldsymbol{x};g(\boldsymbol{t})) & \mbox {if testing,}
\end{cases}
\end{equation}
where the task vector $\boldsymbol{t} \in \mathbb{R}^D$ denotes the control factor, which determines the styles of output images.
For a given image quality measure $\mathcal{P}$, the best task vector ($\boldsymbol{t}^*$) in $M$ styles is formally defined as:
\begin{equation}\label{eq:optimal_task}
\boldsymbol{t}^* \equiv \operatorname*{argmax}_{\boldsymbol{t}_m} \mathcal{P}(ISP(\boldsymbol{x};g(\boldsymbol{t}_m))),
\end{equation}
where $\boldsymbol{t}_m$ is the task vector for the $m$-th style.

\section{Parametric ISP Pipeline}
\label{sec:pipeline}
Recent works on image retouching~\cite{he2020conditional,jie2021LPTN} focus on designing lightweight and fast algorithms using neural networks.
These networks directly transform the input image to its desired form with feature extraction.
However, extracting features from high-resolution images is computationally expensive.
We decompose the role of an image retouching algorithm into image transformation (or ISP pipeline) and the retouching skill prediction.
This decomposition allows the ISP pipeline to have only 19 parameters and the control factor determines retouching skills independently from the input image as presented in Section~\ref{sec:CRISP_scenario}.

This section describes the ISP pipeline, which in order consists of global scaling, white balancing, color space conversion, gamma correction, and tone mapping (See Figure~\ref{fig:framework}{\color{red}(right)}).
Digital cameras have used each function of ISP to transform raw sensor data similar to what the human eye sees~\cite{ramanath2005color}.
By contrast, we use ISP functions for image enhancement transforming flat-looking images to high-quality photographs as in MIT-Adobe FiveK dataset~\cite{fivek}.

\subsection{Digital Gain}
Digital cameras commonly apply a global scaling to all pixel values for intensity adjustment, where the exposure time determines the scaling factor.
By contrast, photographers brighten or dim an image for different image styles.
We assume that the global scaling factor $\phi_{dg}$ is a controllable parameter,
\begin{equation}\label{eq:digital_gain}
\text{gain}(x; \boldsymbol{\phi}) = \phi_{dg} \cdot x,
\end{equation}
where $x$ is a normalized pixel value of an image of which range is [0,1].
The range of $\phi_{dg}$ is [0.85, 2.17] in the test dataset.

\subsection{White Balance}
The color of the illumination changes the color of the objects captured by a camera.
White balance conventionally aims to appear the `true' color of an object or adjust it to the color of different light conditions.
We reinterpret the white balance as a color temperature control by a per-channel scaling function for red and blue colors,
\begin{equation}\label{eq:white_balance}
\text{WB}\left(
\begin{bmatrix}
x_r\\
x_g\\
x_b
\end{bmatrix} ; \boldsymbol{\phi}
\right) = 
\begin{bmatrix}
\phi_{r} \cdot x_r\\
x_g\\
 \phi_{b} \cdot x_b
\end{bmatrix},
\end{equation}
where $x_r$, $x_g$, and $x_b$ are the pixel values for in order red, green, and blue, of which ranges are [0,1].
In the test dataset, $\phi_r$ has a smaller variation of [0.73, 1.07] than $\phi_b$ of [0.80, 2.41].
This result is reasonable since the human visual system is more sensitive to red than blue.

\begin{figure}[t]
	\centering
	\includegraphics[width=1\linewidth]{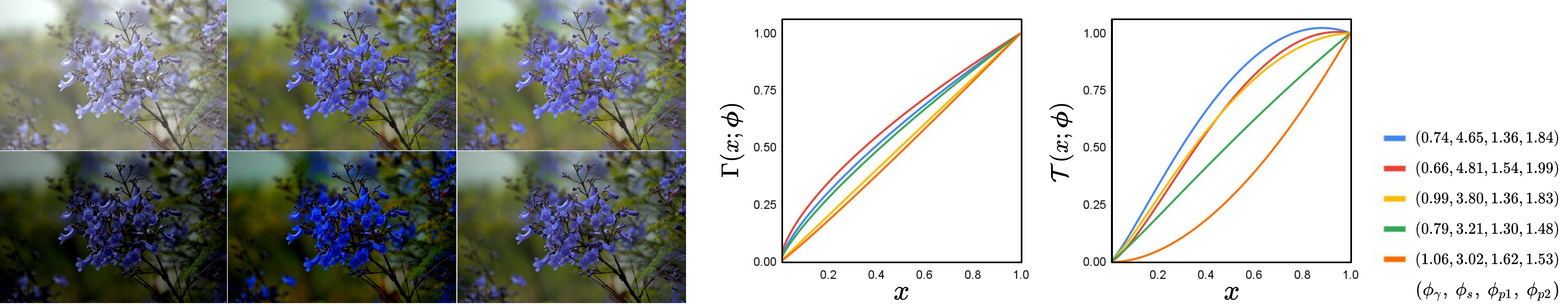}  \\ \vspace{-0.3cm}
	\scriptsize
	\caption{Examples of color Correction matrix (left), gamma correction curve (middle), and tone mapping curve (right).}
	\vspace{-1em}
	\label{fig:ccm}
\end{figure}

\subsection{Color Correction}
In general, the color filters of an image sensor have their own RGB spectra.
Using a color correction matrix (CCM), the image pipeline converts the ``camera space" RGB color to the standard sRGB color.
We assume that the photographer's activity includes the color space conversion.
We adopt the CCM function of which 3$\times$4 parameters are controllable as follows,
\begin{equation}\label{eq:ccm}
\text{CCM}\left(
\begin{bmatrix}
x_r\\
x_g\\
x_b
\end{bmatrix} ; \boldsymbol{\phi}
\right)
=
\begin{bmatrix}
\phi_{11} & \phi_{12} & \phi_{13}\\
\phi_{21} & \phi_{22} & \phi_{23}\\
\phi_{31} & \phi_{32} & \phi_{33}
\end{bmatrix}
\begin{bmatrix}
x_r\\
x_g\\
x_b
\end{bmatrix}
+
\begin{bmatrix}
\phi_{o1}\\
\phi_{o2}\\
\phi_{o3}
\end{bmatrix},
\end{equation}
where $\phi_{oi}$ denotes the color offset for the $i$-th row in the matrix.
We follow a general constraint of CCM as $\sum_j \phi_{ij}=1$ where $i\in\{1,2,3\}$.
Figure~\ref{fig:ccm} presents the color space conversion by the CCM parameters used for the test dataset.

\subsection{Gamma Correction}
The human eye perceives the gradations of color in the dark area better.
Gamma correction is a function that displays the low-intensity pixels with more bits.
While the conventional approach has a fixed function regardless of image contents, we parameterize the gamma correction function:
\begin{equation}\label{eq:digital_gain}
\Gamma(x;\boldsymbol{\phi}) = \text{max}(x,\epsilon)^{\phi_{\gamma}},
\end{equation}
where $\epsilon =10^{-8}$ for the stable training.
By normalizing $x$ to [0,1], gamma correction is an increasing function of which the output range is also [0,1]. 
Figure~\ref{fig:ccm}{\color{red}(left)} displays the gamma curves in the test dataset.

\subsection{Tone Mapping}
When visualizing high dynamic range images (or high-bit images) in 24-bit RGB format, tone mapping curves often adopt S-shape that allocates more bits for mid-intensity values.
Similar to gamma correction, we adopt a simple function that can shape S-curves with parameter tuning,
\begin{equation}\label{eq:digital_gain}
\mathcal{T}(x;\boldsymbol{\phi}) = \phi_{s}\cdot \text{max}(x,\epsilon)^{\phi_{p1}}-(\phi_{s}-1)\cdot \text{max}(x,\epsilon)^{\phi_{p2}},
\end{equation}
where $x$ is the pixel value scaled to [0,1] for a general expression for various bit-widths.
Figure~\ref{fig:ccm}{\color{red}(right)} presents the examples of our tone mapping curves, that pass through (0,0) and (1,1) regardless of the controllable parameters.

\subsection{Parameter Initialization}
Our ISP pipeline consists of simple image transformation functions, originally developed by the characteristic of the human visual system and the image sensor of a digital camera.
Each function has a strong structural prior for raw2rgb conversion, while we adaptively change their parameters for image enhancement.
To avoid indirect solutions of image transformation (\textit{e.g.}, global scaling by a negative value), we initialize the ISP parameters $\boldsymbol{\phi}_{init}$ for similar average pixel values between the outputs and the high-quality images.
Specifically, $\phi_{dg}$ is set to 1.2, WB and CCM set parameters for identity mapping, $\phi_\gamma$ is set to $1\over2.2$, and $\phi_s$, $\phi_{p1}$, and $\phi_{p2}$ are set to 3, 2 and 3, respectively.
In training, our neural network learns the residual of $\boldsymbol{\phi}-\boldsymbol{\phi}_{init}$.

\section{Neural Networks}
We formulate the encoder-decoder framework (See Figure~\ref{fig:framework}{\color{red}(left)}) as neural networks.
Our encoder aims to identify retouching skills used in high-quality images.
To this end, we concatenate low-quality and high-quality images as input for the resnet-style architecture, consisting of 12 convolution layers with 64 channels.
The output of the encoder is the $D$-dimensional non-negative vector of which values determine the styles of output images in test time, described in Equation~\ref{eq:isp}.
Our decoder consists of 5 fully connected layers with 64 channels that predict the residuals to initial ISP parameters $\boldsymbol{\phi}_{init}$.
The decoder is computationally efficient since it has a spatial resolution of 1$\times$1 and can be computed independently to the test image. 
When the task vector (or the decoder's input) is a zero vector, the ISP parameters are $\boldsymbol{\phi}_{init}$.
The network architecture details are described in the supplementary document.

\section{Experiments}
\label{sec:experiments}

\subsection{Dataset}
We train and evaluate our method on the MIT-Adobe FiveK dataset~\cite{fivek} that consists of 5,000 camera raw images and paired RGB images retouched by five experts, denoted as Expert-A, Expert-B, Expert-C, Expert-D, and Expert-E.
State-of-the-art methods on this dataset commonly downscale raw and RGB images and preprocess raw images to low-quality RGB images.
For the fair comparisons, we conduct two datasets following the downscaling and preprocessing settings in CSRNet~\cite{he2020conditional} using Lightroom\footnote{https://github.com/yuanming-hu/exposure/wiki/Preparing-data-for-the-MITAdobe-FiveK-Dataset-with-Lightroom} and 3DLUT~\cite{zeng2020learning} by downloading the released dataset\footnote{https://github.com/HuiZeng/Image-Adaptive-3DLUT}.
The dataset for CSRNet downscaled images to 500 pixels on the long edge, while the dataset for 3DLUT resized images to 480 pixels on the short edge.
Each dataset has different lists of test images, of which numbers are 500 and 498.
For fair comparisons, we use the remaining images from Expert-C in each dataset for training unless otherwise specified.

\subsection{Implementation Details}
We set the task vector as three controllable parameters ($D=3$).
We adopt an MSE loss between outputs and images from Expert-C to train encoder and decoder networks with Adam optimizer~\cite{kingma2015adam}.
We randomly crop images to 200$\times$200 pixels and flip and rotate the patches for augmentation.
We set the batch size to 16 and the initial learning rate to  $1\times 10^{-4}$.
During $1.6\times 10^{5}$ iterations, we halve the learning rate for every quarter of training.
We use the image quality measures ($\mathcal{P}$) as MOS, PSNR (dB), and SSIM, and evaluate the model efficiency with model parameters, floating-point operations (FLOPs), and the number of inferences to find desired results.
It is challenging to measure how easily users can control a system. 
We use a simple greedy search algorithm (See Algorithm~\ref{alg:greedy}) as a proxy of user behavior in finding desired results to evaluate the proxy in quantitative measures.
Initial task vector ($\boldsymbol{t}_{init}$), step size ($s$), and stop condition ($K$) represent the propensity of user behaviors (meticulous or hasty) that affect the output image quality and the number of inferences.
We use Algorithm~\ref{alg:greedy} for reference-based image quality measures (PSNR/SSIM), where $\boldsymbol{t}_{init}$, $s$, and $K$ are (0.0,0.0,0.0), 0.1, and 100, representing a meticulous user to achieve high image quality, unless otherwise specified.
$t_d$ denotes the $d$-th element of $\boldsymbol{t}$. 

\begin{algorithm}
\footnotesize
\caption{\footnotesize Greedy Search Algorithm}\label{alg:greedy}
\begin{algorithmic}
\Require Initial task vector $\boldsymbol{t}_{init} \in \mathbb{R}^D$, Step size $s$, Stop condition $K$
\Require Input image $\boldsymbol{x}$, Reference image $\hat{\boldsymbol{x}}$
\State $e \gets \inf,~ \boldsymbol{t} \gets \boldsymbol{t}_{init},~ d \gets 0,~ i \gets 0,~ k \gets 0$
\While{$k \leq K$}
\If{$e < \text{MSE}(ISP(\boldsymbol{x};g(\boldsymbol{t})), \hat{x})$}
    \State $t_{d+1} \gets t_{d+1} - s,~ d \gets (d+1)\%D,~ i \gets i+1,~ k \gets k+1$
    \If{$i = D$}
        \State $\boldsymbol{t} \gets \boldsymbol{t}+s,~ i\gets0 $ 
    \EndIf
\Else
\State $e \gets \text{MSE}(ISP(\boldsymbol{x};g(\boldsymbol{t})), \hat{x}),~ i \gets 0,~ k \gets 0$
\EndIf
\State $t_{d+1} \gets t_{d+1} + s$
\EndWhile
\end{algorithmic}
\end{algorithm}
\vspace{-1em}

\begin{figure*}[h!]
	\centering\scriptsize
    \scalebox{1}{
	\setlength\tabcolsep{1pt}
    \begin{tabular}{ccccc} 
     \vspace{-0.5cm}\\
\includegraphics[trim= 0 140 0 60, clip ,width=0.195\linewidth]{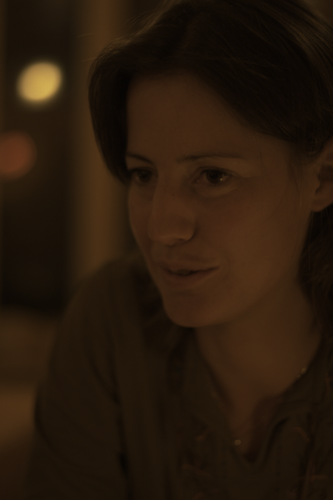} & 
\includegraphics[trim=0 140 0 60, clip,width=0.195\linewidth]{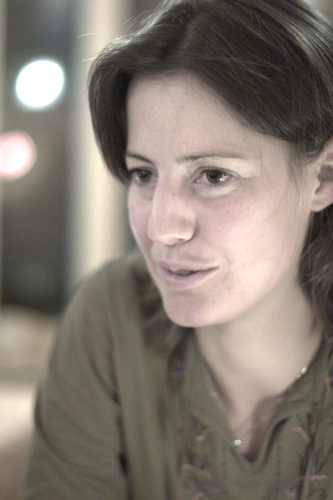} & 
\includegraphics[trim=0 140 0 60, clip,width=0.195\linewidth]{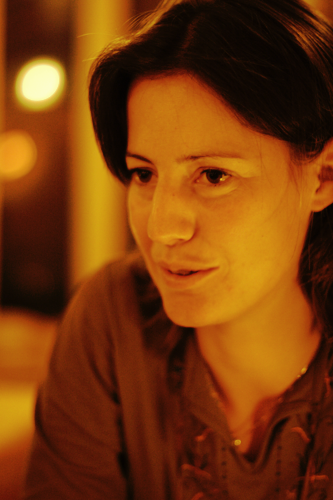} & 
\includegraphics[trim=0 140 0 60, clip,width=0.195\linewidth]{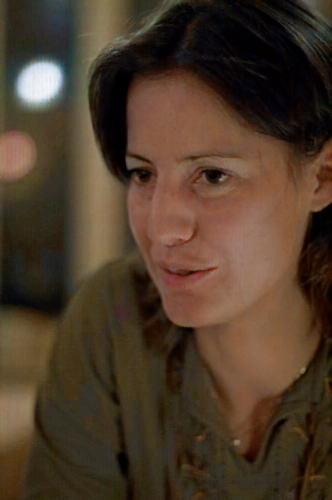} & \includegraphics[trim=0 140 0 60, clip,width=0.195\linewidth]{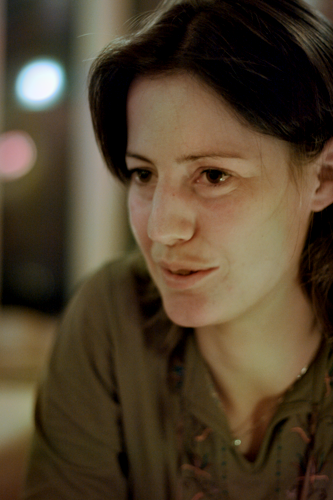} \vspace{-0.05cm}\\
Input & DAR~\cite{park2018distort} & White-Box~\cite{hu2018exposure} & Pix2Pix~\cite{isola2017pix2pix} &  HDRNet~\cite{gharbi2017deep} \\

\includegraphics[trim=0 140 0 60, clip,width=0.195\linewidth]{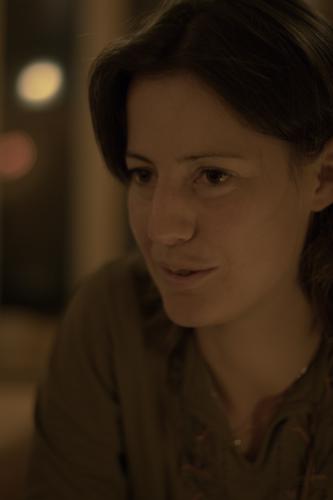} & 
\includegraphics[trim=0 140 0 60, clip,width=0.195\linewidth]{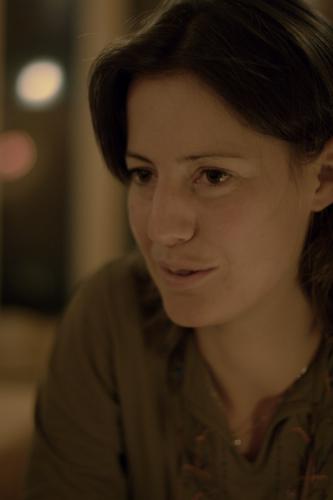} &
\includegraphics[trim=0 140 0 60, clip,width=0.195\linewidth]{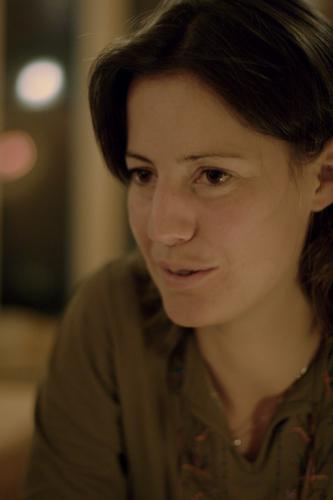} &
\includegraphics[trim=0 140 0 60, clip,width=0.195\linewidth]{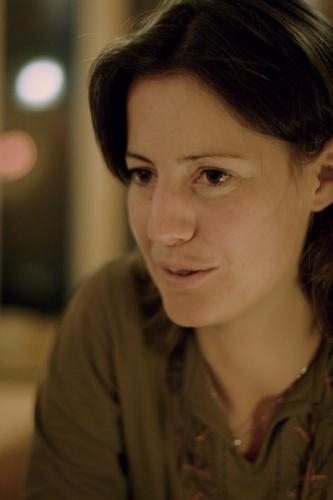} &
\includegraphics[trim=0 140 0 60, clip,width=0.195\linewidth]{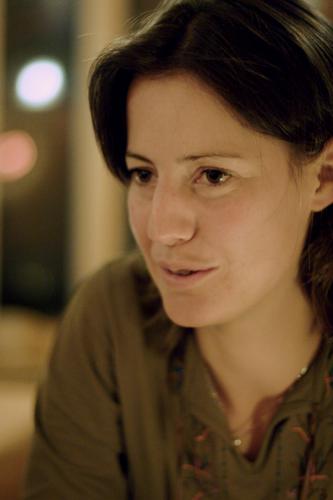} \vspace{-0.05cm}\\
 \multicolumn{5}{c}{$\xleftarrow{\hspace*{5.1cm}}$ CSRNet~\cite{he2020conditional} $\xrightarrow{\hspace*{5.1cm}}$} \\
\includegraphics[trim=0 140 0 60, clip, width=0.195\linewidth]{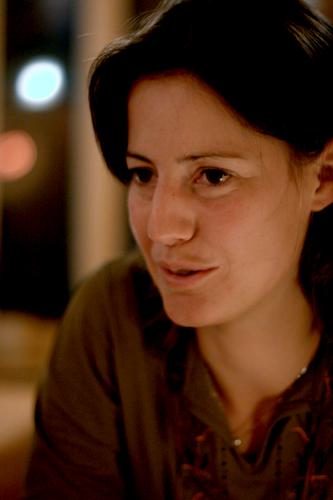} & \includegraphics[trim=0 140 0 60, clip, width=0.195\linewidth]{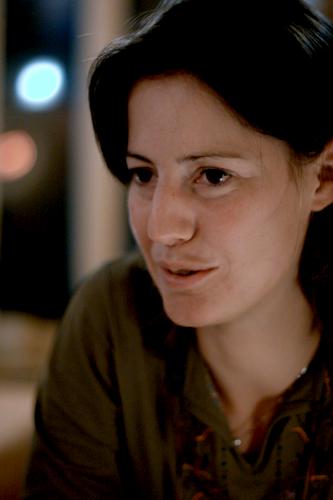} & \includegraphics[trim=0 140 0 60, clip, width=0.195\linewidth]{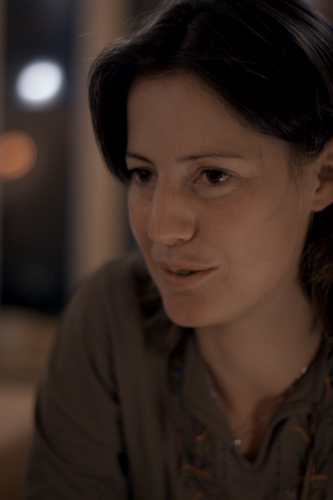} & \includegraphics[trim=0 140 0 60, clip, width=0.195\linewidth]{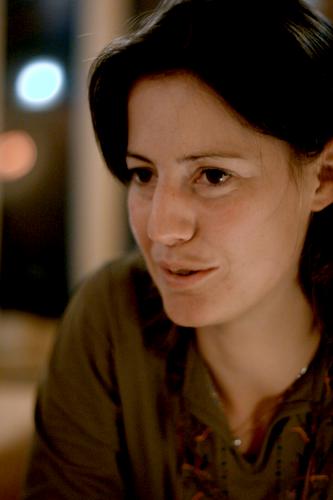} & \includegraphics[trim=0 140 0 60, clip, width=0.195\linewidth]{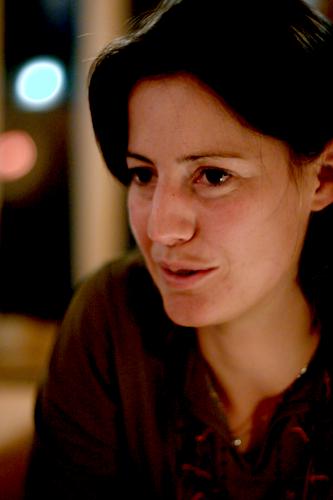} \vspace{-0.05cm}\\
 \multicolumn{5}{c}{$\xleftarrow{\hspace*{4.9cm}}$ CRISP (Ours) $\xrightarrow{\hspace*{4.9cm}}$}\\
 \includegraphics[trim=0 140 0 60, clip, width=0.195\linewidth]{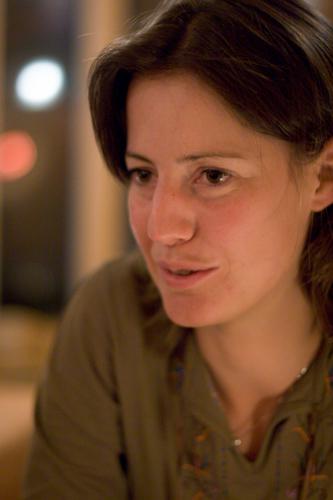} &
 \includegraphics[trim=0 140 0 60, clip, width=0.195\linewidth]{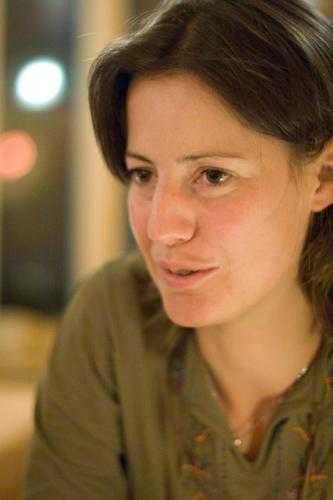} &
 \includegraphics[trim=0 140 0 60, clip, width=0.195\linewidth]{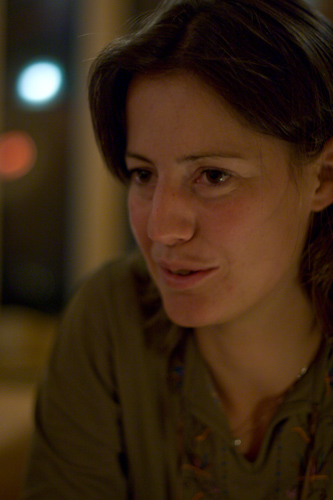} &
\includegraphics[trim=0 140 0 60, clip, width=0.195\linewidth]{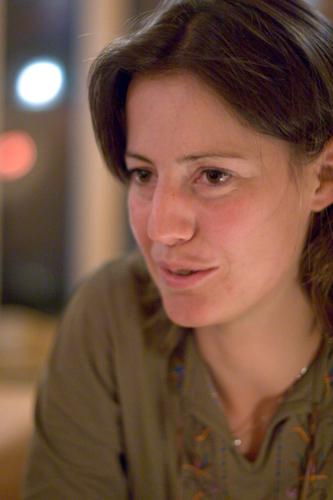} &
\includegraphics[trim=0 140 0 60, clip, width=0.195\linewidth]{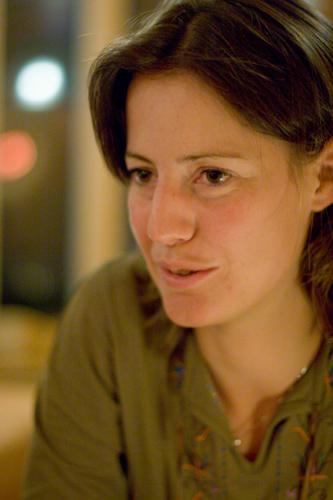}  \vspace{-0.05cm}\\
 \multicolumn{5}{c}{$\xleftarrow{\hspace*{5.15cm}}$ Expert~\cite{fivek} $\xrightarrow{\hspace*{5.15cm}}$}\\

\includegraphics[width=0.195\linewidth]{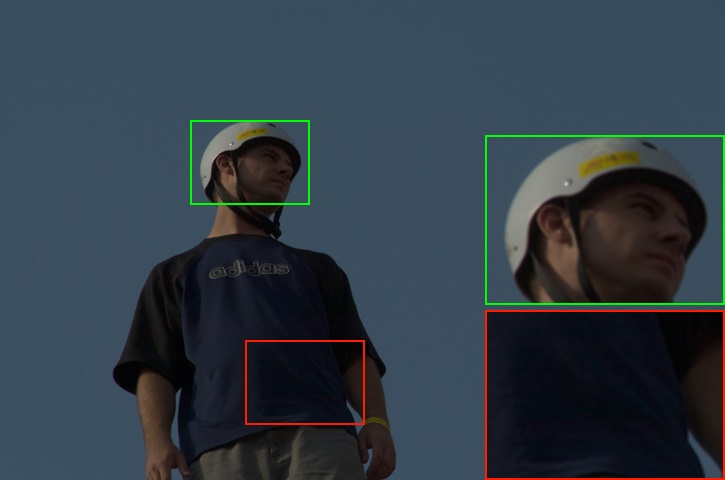} & 
\includegraphics[width=0.195\linewidth]{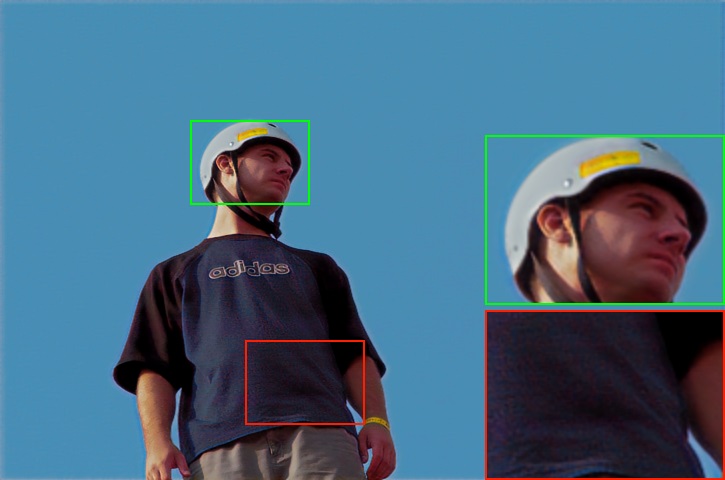} & 
\includegraphics[width=0.195\linewidth]{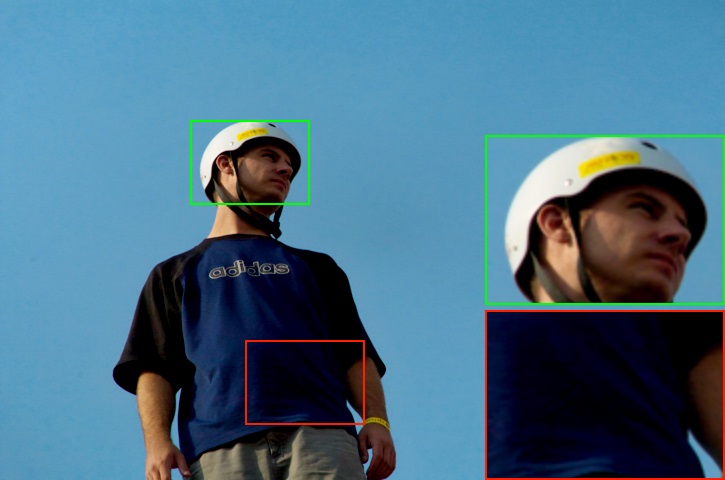} & \includegraphics[width=0.195\linewidth]{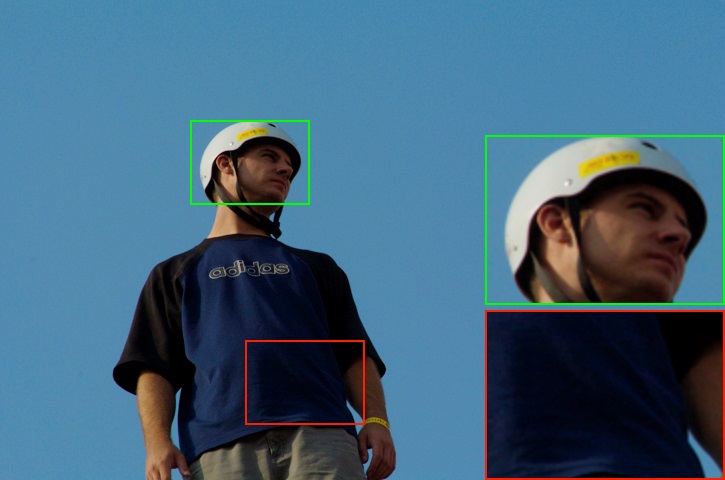} & \includegraphics[width=0.195\linewidth]{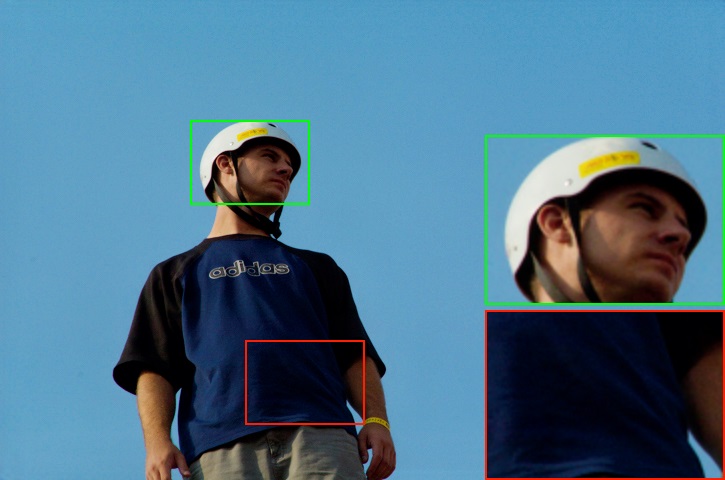} \vspace{-0.05cm}\\
Input & DPED~\cite{ignatov2017dslr} & DeepLPF~\cite{moran2020depplpf} & 3DLUT~\cite{zeng2020learning} & SA3DLUT~\cite{wang2021realtime}\\
\includegraphics[width=0.195\linewidth]{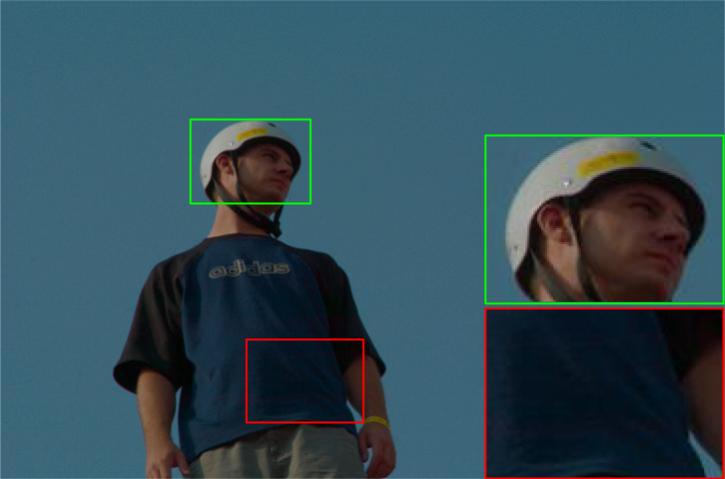} &
\includegraphics[width=0.195\linewidth]{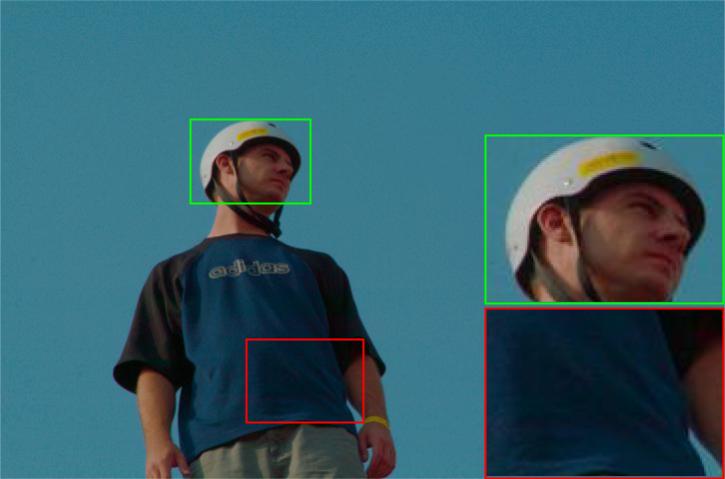} &
\includegraphics[width=0.195\linewidth]{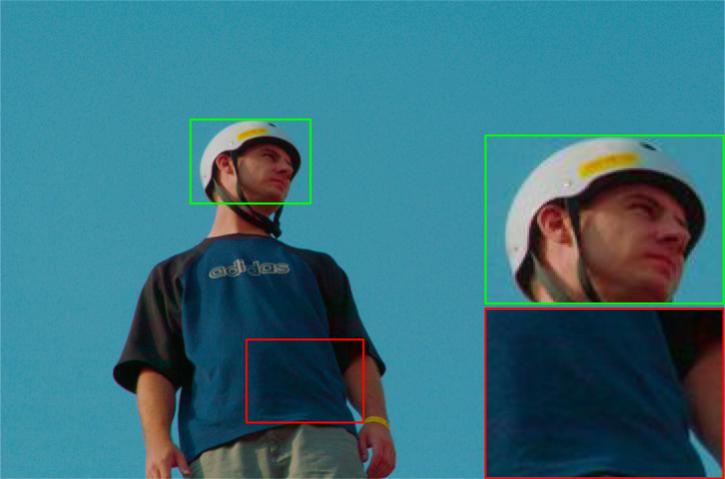} &
\includegraphics[width=0.195\linewidth]{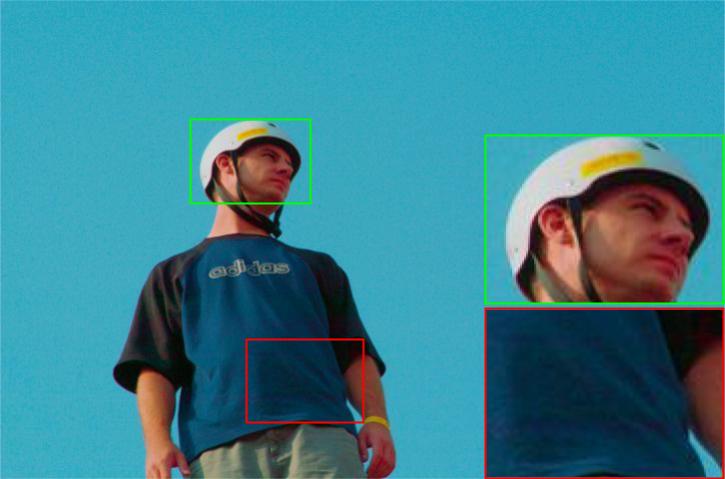} &
\includegraphics[width=0.195\linewidth]{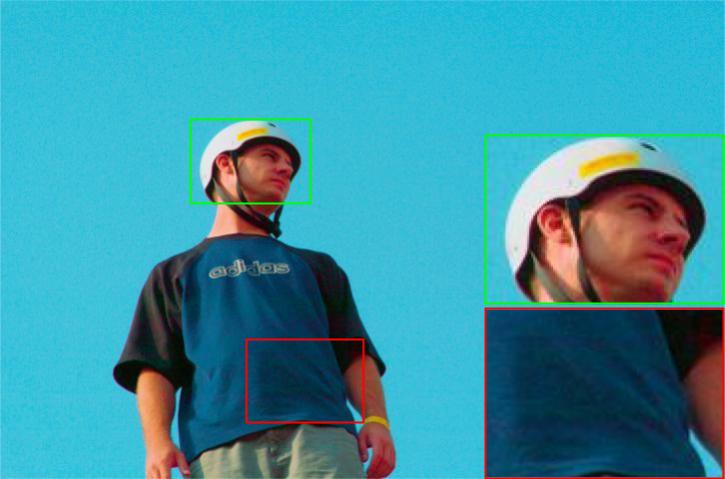} \vspace{-0.05cm}\\
 \multicolumn{5}{c}{$\xleftarrow{\hspace*{5.1cm}}$ CSRNet~\cite{he2020conditional} $\xrightarrow{\hspace*{5.1cm}}$} \\
\includegraphics[width=0.195\linewidth]{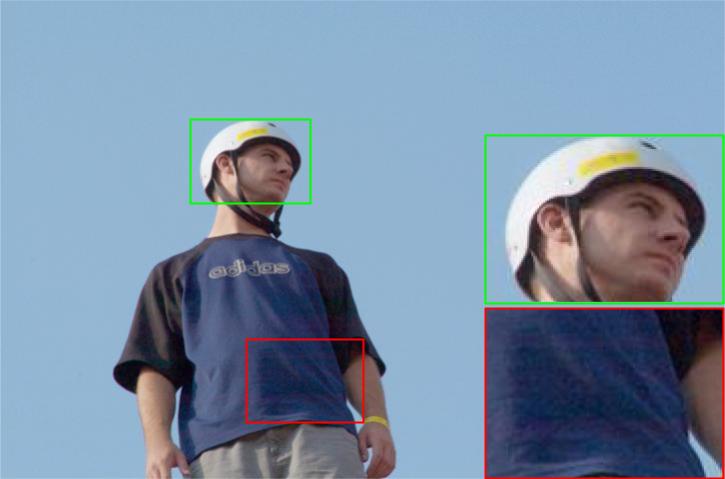} & 
\includegraphics[width=0.195\linewidth]{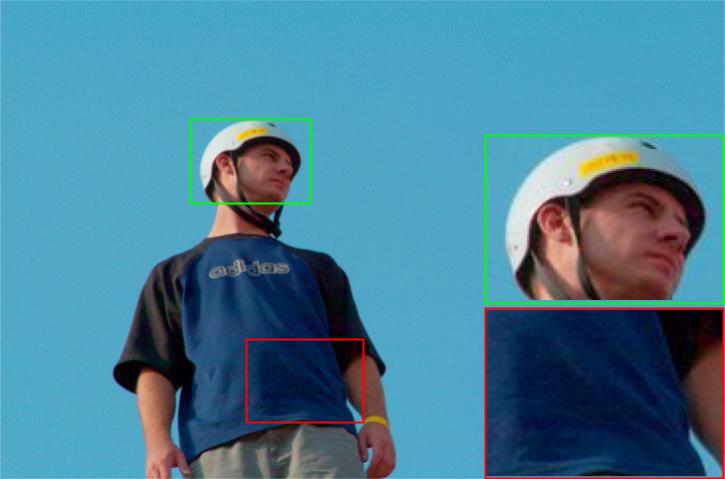} & 
\includegraphics[width=0.195\linewidth]{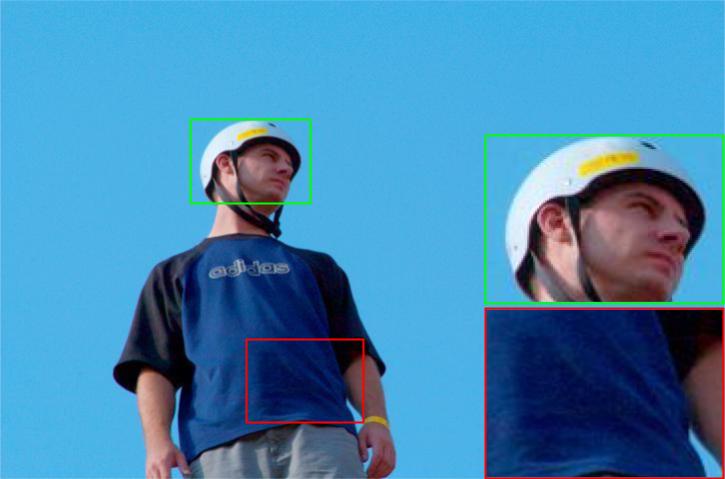} & 
\includegraphics[width=0.195\linewidth]{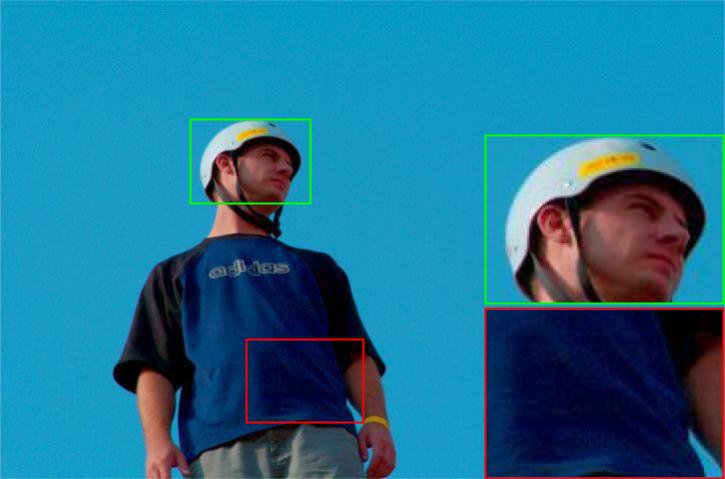} & 
\includegraphics[width=0.195\linewidth]{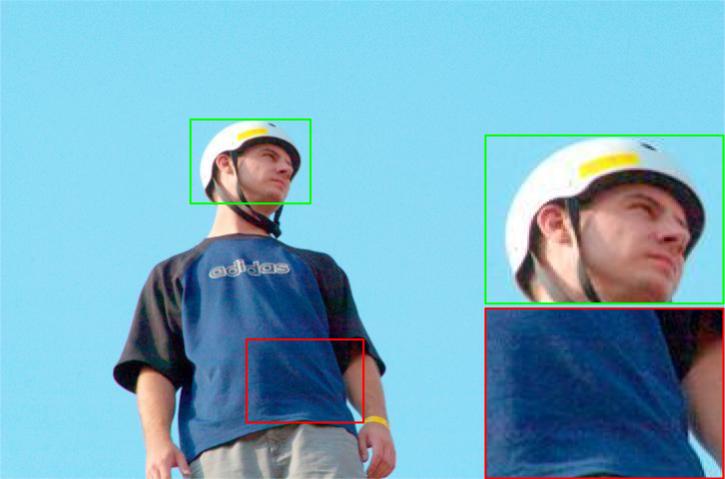} 
\vspace{-0.05cm}\\
 \multicolumn{5}{c}{$\xleftarrow{\hspace*{4.9cm}}$ CRISP (Ours) $\xrightarrow{\hspace*{4.9cm}}$}\\
\includegraphics[width=0.195\linewidth]{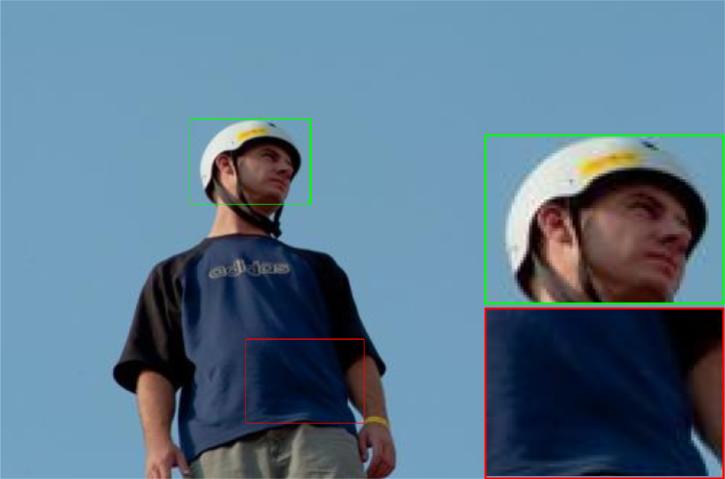} & 
\includegraphics[width=0.195\linewidth]{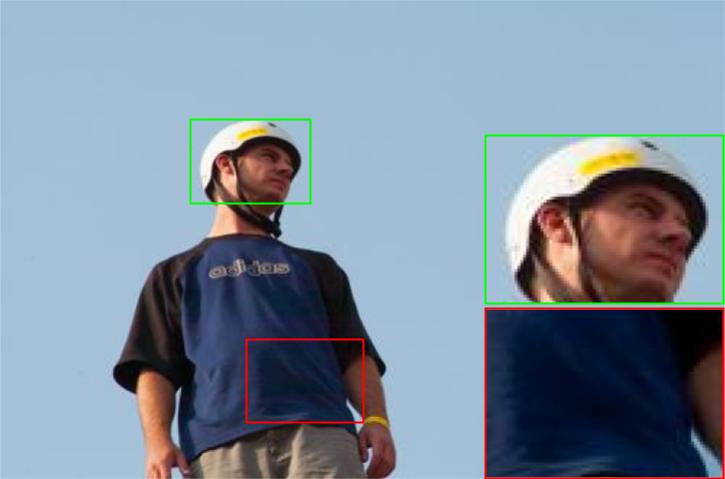} & 
\includegraphics[width=0.195\linewidth]{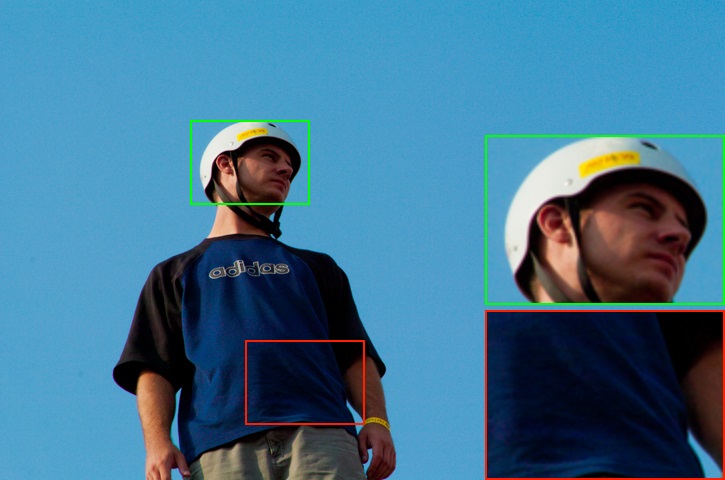} & 
\includegraphics[width=0.195\linewidth]{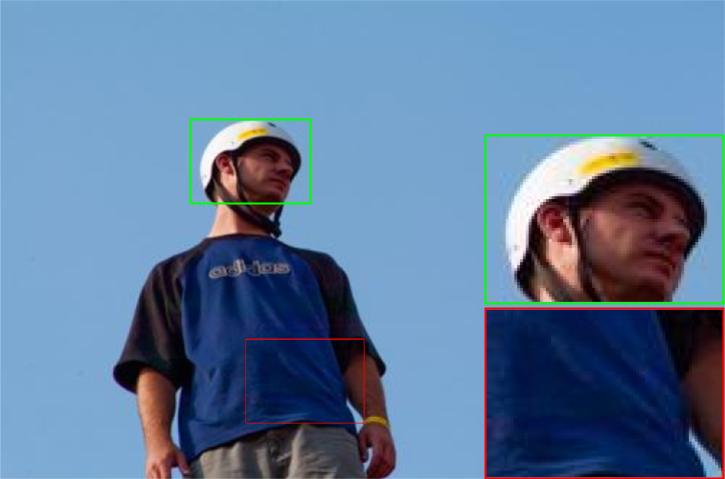} & 
\includegraphics[width=0.195\linewidth]{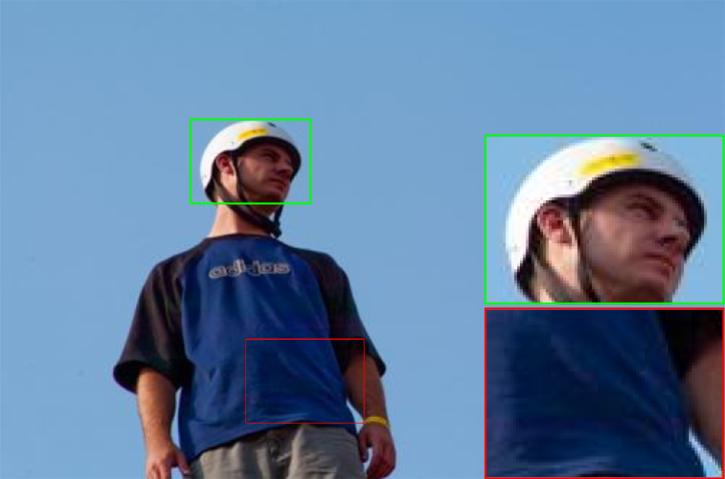} \vspace{-0.05cm}\\
 \multicolumn{5}{c}{$\xleftarrow{\hspace*{5.15cm}}$ Expert~\cite{fivek} $\xrightarrow{\hspace*{5.15cm}}$}
	\end{tabular}} \vspace{-0.4cm}
	\caption{Qualitative comparison with SotA models. This figure visualizes five enhanced images with different styles for our method (CRISP), CSRNet, and Expert. The single enhanced image for each input is visualized for the other approaches. The outputs of CRISP are natural and vivid with a variety of styles compared to all other methods.}
	\vspace{-1em}
	\label{fig:sota}
\end{figure*}

\subsection{Comparisons with SotA methods}

We compare our method with 11 state-of-the-art methods, DAR~\cite{park2018distort}, White-Box~\cite{hu2018exposure}, Pix2Pix~\cite{isola2017pix2pix}, HDRNet~\cite{gharbi2017deep}, DPED~\cite{ignatov2017dslr}, DeepLPF~\cite{moran2020depplpf}, 3DLUT~\cite{zeng2020learning}, SA3DLUT~\cite{wang2021realtime}, and CSRNet~\cite{he2020conditional}.
Expert~\cite{fivek} denotes the images enhanced by human experts in the MIT-Adobe FiveK dataset~\cite{fivek}.

\textbf{Visual comparison.}
Figure~\ref{fig:sota} visualizes the results of the state-of-the-art methods.
The input images from the MIT-Adobe FiveK dataset generally have low pixel values to avoid saturation.
DAR, White-Box, and HDRNet brighten the input image but change its natural color.
Pix2Pix generates a natural colored image but it contains checkerboard artifacts.
DPED, DeepLPF, 3DLUT, and SA3DLUT output realistic images but their colors are generally dark.
CSRNet generates a bright image with natural tone (the right most images in Figure~\ref{fig:sota}) and allows intermediate image generation between the input and the output.
Intermediate images differ in image intensity, but are similar in tone and color.
By contrast, our method (CRISP) generates diverse realistic styles with natural and vivid colors like the human experts (Expert) in the dataset~\cite{fivek}.
Please see the supplementary document for more comparisons.

\begin{figure}[t]
	\centering\scriptsize
	\hspace{-0.6cm}
    \scalebox{1}{
	\setlength\tabcolsep{6pt}
    \begin{tabular}{ccc}
\includegraphics[width=0.4\linewidth]{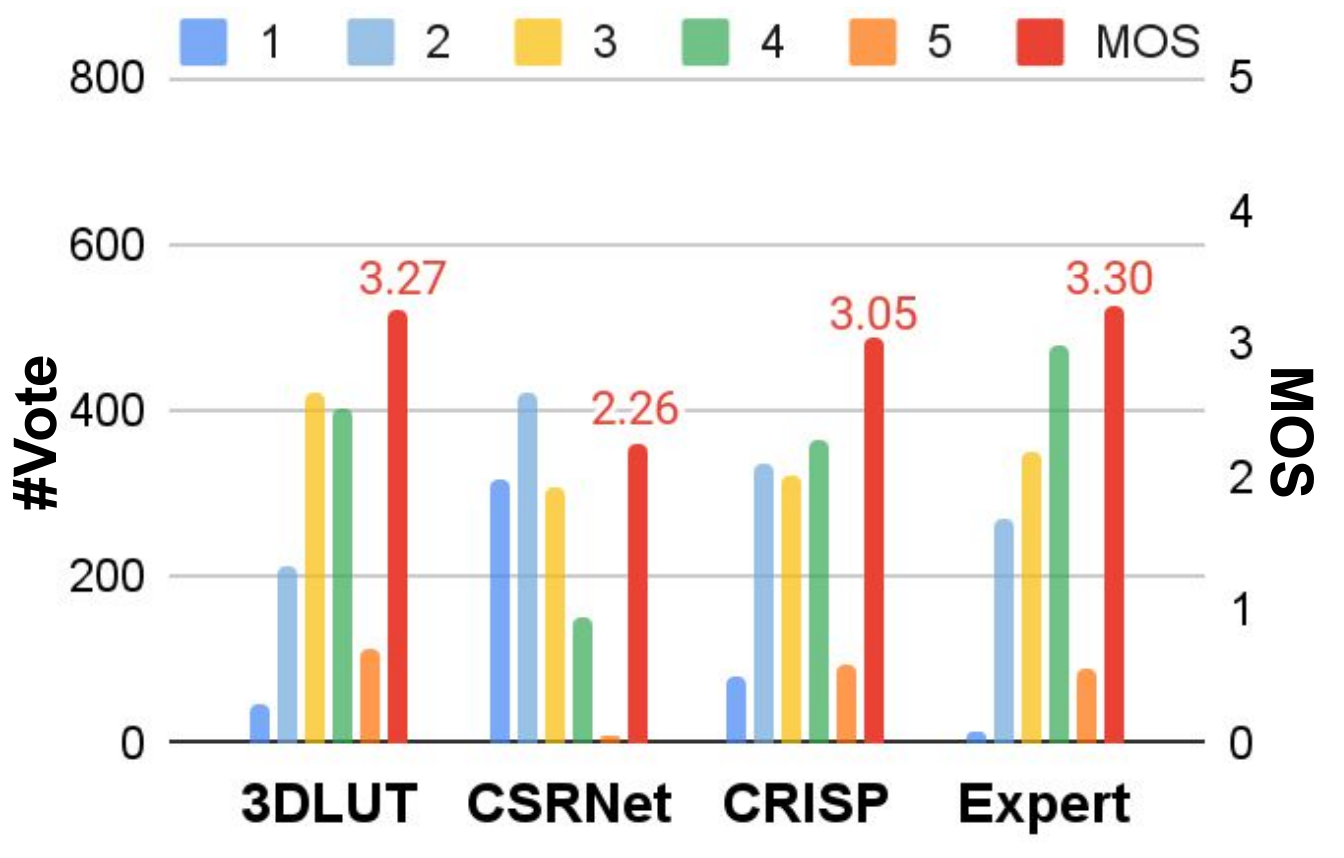} & 
\includegraphics[width=0.4\linewidth]{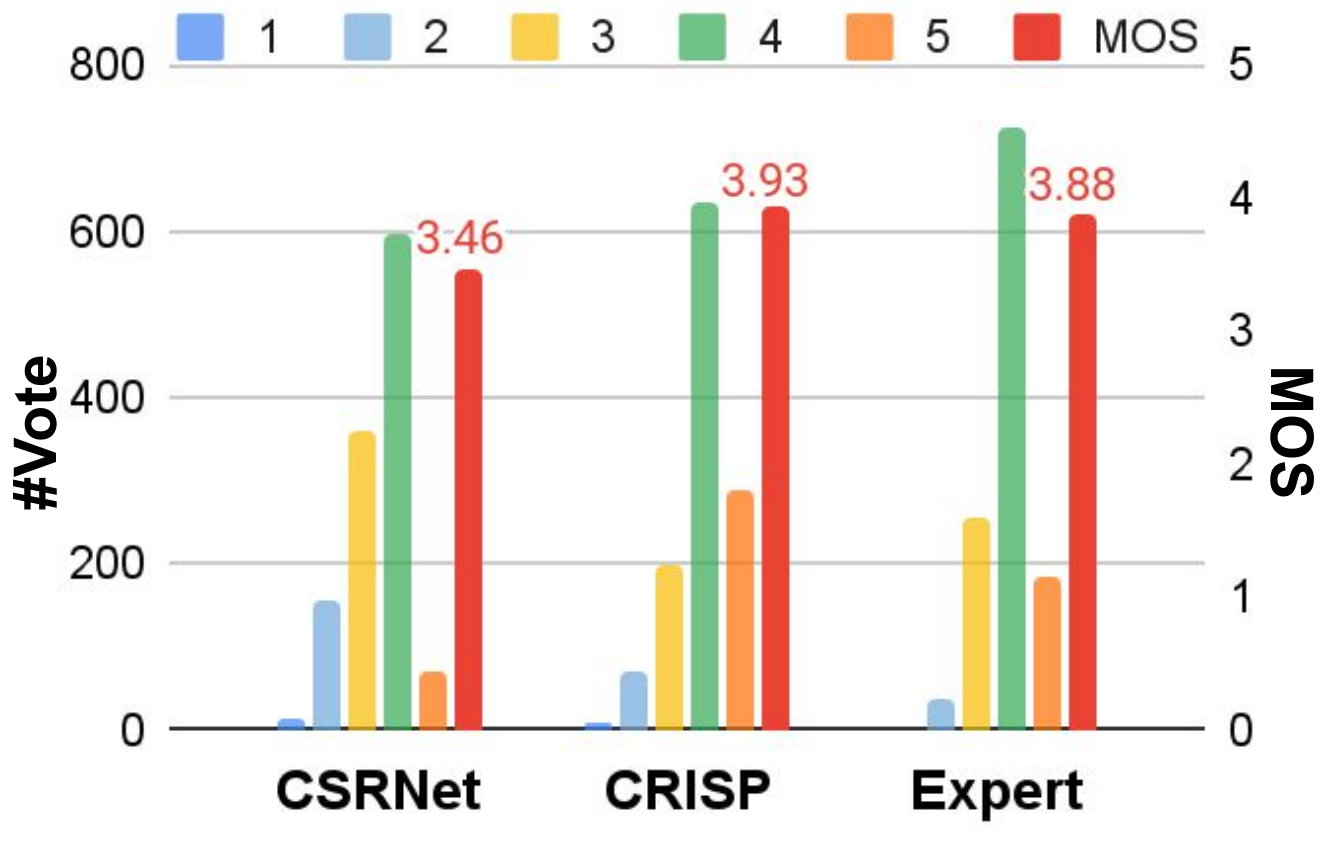}  \vspace{-0.1cm} \\
(a) Single-style MOS & (b) Multiple-style MOS
	\end{tabular}} \vspace{-0.3cm}
	\scriptsize
	\captionof{figure}{User study results of mean opinion score (MOS) on MIT-Adobe FiveK. For each method 1200 samples (40 images $\times$ 30 participants) were assessed. MOS scores (red blocks) are along the right axis and the numbers of votes for each rating (the other color blocks) are along the left axis. (a) Participants rate the image quality of a single enhanced output for each input. (b) Participants rate the image quality of the output they prefer the most out of five styles for each input. CRISP in (b) outperforms all other methods including Expert.}
	\label{fig:mos}
	\vspace{-1em}
\end{figure}

\textbf{User study.}
We conducted two types of user study with 30 participants and 40 low-quality images to quantify the effectiveness of multiple style generation for image enhancement.
The participants are asked to rate the image quality from 1 (bad quality) to 5 (excellent quality) to enhanced images.
In the first type of user study, called single-style MOS (Mean Opinion Score), the participants rate the quality of the single-style outputs from 3DLUT, CSRNet, CRISP, and Expert.
We use the outputs from 3DLUT and CSRNet and randomly select an output from CRISP and Expert for each input as a style.
In the second type of user study, called multiple-style MOS, the participants first select the most preferred image among five different styles from CSRNet, CRISP, and Expert and then rate the quality of the selected images.
We use intermediate images between the input and the output for CSRNet, randomly selected images for CRISP, and images enhanced by five human photographers for Expert (See Figure~\ref{fig:sota}) as different styles.
Figure~\ref{fig:mos} visualizes the results of the user studies.
In single-style MOS, CSRNet has the lowest score since it often generates too bright images.
All methods including Expert often fail to satisfy user preferences as rating scores of 1 and 2.
By contrast, multiple-style MOS results of CSRNet, CRISP, and Expert outperform their scores for single-style MOS by a large margin.
CRISP has the largest number of votes for 5 (excellent quality) and the highest MOS compared to all other methods including Expert.
The average number of images that were not selected by participants was 0.8 out of 5, indicating various user preferences.
Please see the supplementary document for more images used in user studies.

\textbf{Ability to generate multiple styles.}
Although CRISP outperforms CSRNet in the user study of multiple styles (See Figure~\ref{fig:mos}(b)), it is still a question of how versatile CRISP can generate HQ images.
The images retouched by different experts in the MIT-Adobe FiveK dataset have distinct retouching styles.
To quantify the ability of multiple style generation, we find the outputs most similar to the images retouched by each expert and measure PSNR between them.
While CSRNet changes the image intensity for different styles, CRISP adjusts the tone and color (See Figure~\ref{fig:generalization}(left)).
CRISP outperforms CSRNet by over 4 dB for all experts, as described in Figure~\ref{fig:generalization}(right).
For CSRNet, we generate 100 intermediate images and choose the image with the highest PSNR for each expert.
For CRISP, we use Algorithm~\ref{alg:greedy} for each expert.

\begin{table}[t]

\begin{minipage}[c]{0.5\textwidth}%
	\centering\scriptsize
    \scalebox{1}{
	\setlength\tabcolsep{1pt}
    \begin{tabular}{cccccc}
    Expert & CSRNet & CRISP \\
    \includegraphics[width=0.3\linewidth]{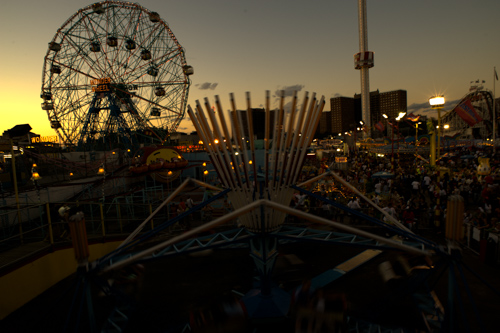} & \includegraphics[width=0.3\linewidth]{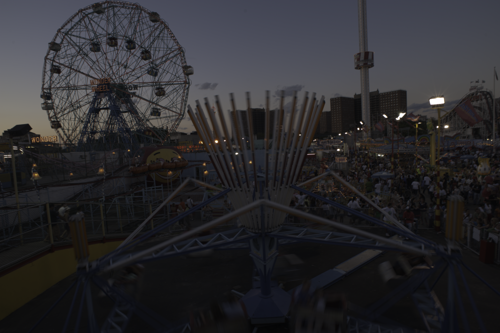} &  \includegraphics[width=0.3\linewidth]{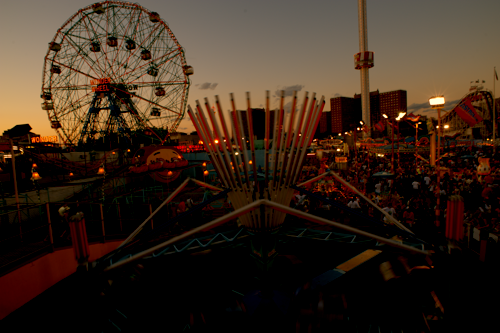} \\
    \includegraphics[width=0.3\linewidth]{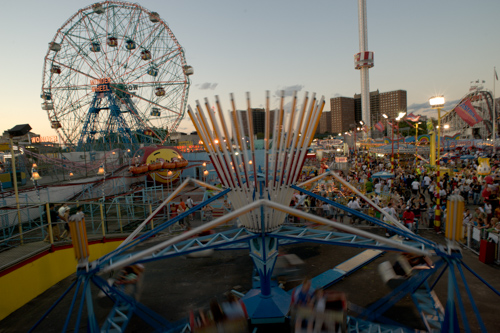} & \includegraphics[width=0.3\linewidth]{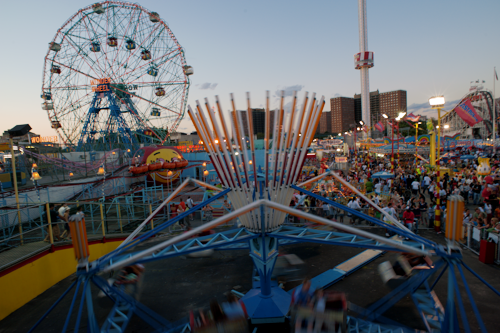} & \includegraphics[width=0.3\linewidth]{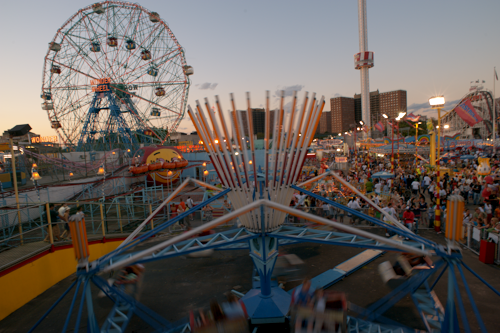} \\
	\end{tabular}} 
	\scriptsize
\end{minipage}
~~
\begin{minipage}[c]{0.4\textwidth}%
    	\centering\scriptsize
    	\scalebox{1}{
    	\setlength\tabcolsep{2pt}
        	\begin{tabular}{lccccccc} 
                \toprule[1.5pt]
                PSNR (dB) & CSRNet & CRISP \\
                \hline \\[-8pt]
                Expert-A & 24.15 &  28.29 \\
                Expert-B & 24.95 & 30.60 \\
                Expert-C & 24.98 & 29.61 \\
                Expert-D & 22.73 & 29.28 \\
                Expert-E & 21.74 & 29.69 \\
                \bottomrule[1.5pt]
        	\end{tabular}}
\end{minipage}
\captionof{figure}{Performances on multiple style generation. (left) Quantitative results of the most similar output images to the images from Expert. (right) PSNR results to the images retouched by different experts.}
\label{fig:generalization}
\vspace{-1em}
\end{table}

\textbf{Efficiency comparison.}
We compare CRISP with SotA methods in PSNR, SSIM, model parameters, FLOPs, and the number of inferences to find desired results.
Specifically, for two widely used benchmark datasets on the MIT-Adobe FiveK dataset with the images from Expert-C, we replicate PSNR and SSIM scores of compared methods from CSRNet~\cite{he2020conditional} and SA3DLUT~\cite{wang2021realtime}.
We reproduce missing scores in two papers for fair comparisons.
We measure FLOPs with the average of a single inference on each dataset.
We use Algorithm~\ref{alg:greedy} with different hyper-parameters to evaluate the number of inferences to find the desired result.
CSRNet, 3DLUT, and SA3DLUT, the most recent efficient models, present the main comparisons with CRISP by the following characteristics:
\begin{itemize}\setlength\itemsep{-0.1em}
    \item \textbf{CSRNet~\cite{he2020conditional}} adaptively modulates features depending on input low-quality images.
    It adopts only fully connected layers for image processing, while convolution layers predict affine transformation parameters for the modulation.
    \item \textbf{3DLUT~\cite{zeng2020learning}} uses input-adaptive trilinear interpolation for image processing. 
    Trilinear interpolation allows FLOPs-efficient inference but requires a relatively large number of parameters for lookup tables.
    \item \textbf{SA3DLUT~\cite{wang2021realtime}} adopts spatially adaptive 3DLUT that requires more FLOPs and parameters for spatially varying image adjustment.
    \item \textbf{CRISP (Ours)} uses simple arithmetic functions for image processing and fully connected layers to adjust the parameters of the arithmetic functions.
\end{itemize}
Table~\ref{table:sota} presents that CRISP has two times smaller parameters and 160 times reduced FLOPs than CSRNet since our model does not use convolution or fully connected layers for image processing.
CRISP requires less than 100 FLOPs for individual pixels, which is less than FLOPs for trilinear interpolation in 3DLUT and SA3DLUT.
Figure~\ref{fig:greedysearch} demonstrates the trade-off between the output image quality (PSNR) and the number of inferences, where each blue dot represents a proxy of a distinct user behavior.
CRISP outperforms PSNR of 3DLUT with five inferences requiring fewer FLOPs than a single inference for 3DLUT while $t_{init}$, $s$, and $K$ being (3,3,3), 3, and 4.

\begin{table}[t]

\begin{minipage}[c]{0.62\textwidth}%
\centering\scriptsize
		\captionof{table}{Quantitative comparison with the SotA methods on MIT-Adobe FiveK (Expert-C).}
    	\label{table:sota}
    	\vspace{-0.0cm}\hspace{-0.3cm}
    	\scalebox{1}{
    	\setlength\tabcolsep{4pt}
        	\begin{tabular}{lccrr} 
            	\toprule[1.5pt]
            	Method & PSNR & SSIM & Params & FLOPs\\
        		\midrule[1.0pt]
            	White-Box~\cite{hu2018exposure} & 18.59 & 0.797 &  8.56$\times 10^6$ & -\\
            	DAR~\cite{park2018distort} & 19.54 & 0.800 & 2.59$\times 10^8$ & - \\
            	HDRNet~\cite{gharbi2017deep} &  22.65 & 0.880 &  4.82$\times 10^5$ & -\\
            	Pix2Pix~\cite{isola2017pix2pix}& 22.05 & 0.788 & 1.14$\times 10^7$ & 5.68$\times 10^{10}$\\
            	CSRNet~\cite{he2020conditional}  &  23.69 & 0.895 & 3.65$\times 10^4$ & 2.17$\times 10^9$\\
                 \hline \\[-6pt]
            	\textbf{CRISP (Ours)} & \textbf{29.61} & \textbf{0.920} & \textbf{1.37}$\times \boldsymbol{10^4}$ & \textbf{1.29}$\times \boldsymbol{10^7}$ \\
        		\midrule[1.0pt]
                DPE~\cite{chen2018dpe} &  23.76 & 0.881 &  3.34$\times 10^6$ & -\\
                DPED~\cite{ignatov2017dslr} &  24.06 & 0.856 & - & -\\
                LPTN~\cite{jie2021LPTN} & 22.14 & 0.854 & 4.03$\times 10^5$ & 5.08$\times 10^9$\\
                CSRNet~\cite{he2020conditional} &  24.23 & 0.900 & 3.65$\times 10^4$ & 4.49$\times 10^9$\\
                DeepLPF~\cite{moran2020depplpf} & 25.29 & 0.899 & 8.00$\times 10^8$ & - \\
                3DLUT~\cite{zeng2020learning} & 25.24 & 0.886 & 5.93$\times 10^5$ & 2.06$\times 10^8$\\
                SA3DLUT~\cite{wang2021realtime} & 25.50 & 0.890 & 4.52$\times 10^6$ & 1.11$\times 10^9$\\
                 \hline \\[-6pt]
            	\textbf{CRISP (Ours)} & \textbf{30.99} & \textbf{0.924} & \textbf{1.37}$\times \boldsymbol{10^4}$ & \textbf{2.66}$\times \boldsymbol{10^7}$\\
                \bottomrule[1.5pt]
        	\end{tabular}} 
\end{minipage}
~~
\begin{minipage}[c]{0.35\textwidth}%
	\centering\scriptsize
	\vspace{0.4cm}
	\includegraphics[width=1\linewidth]{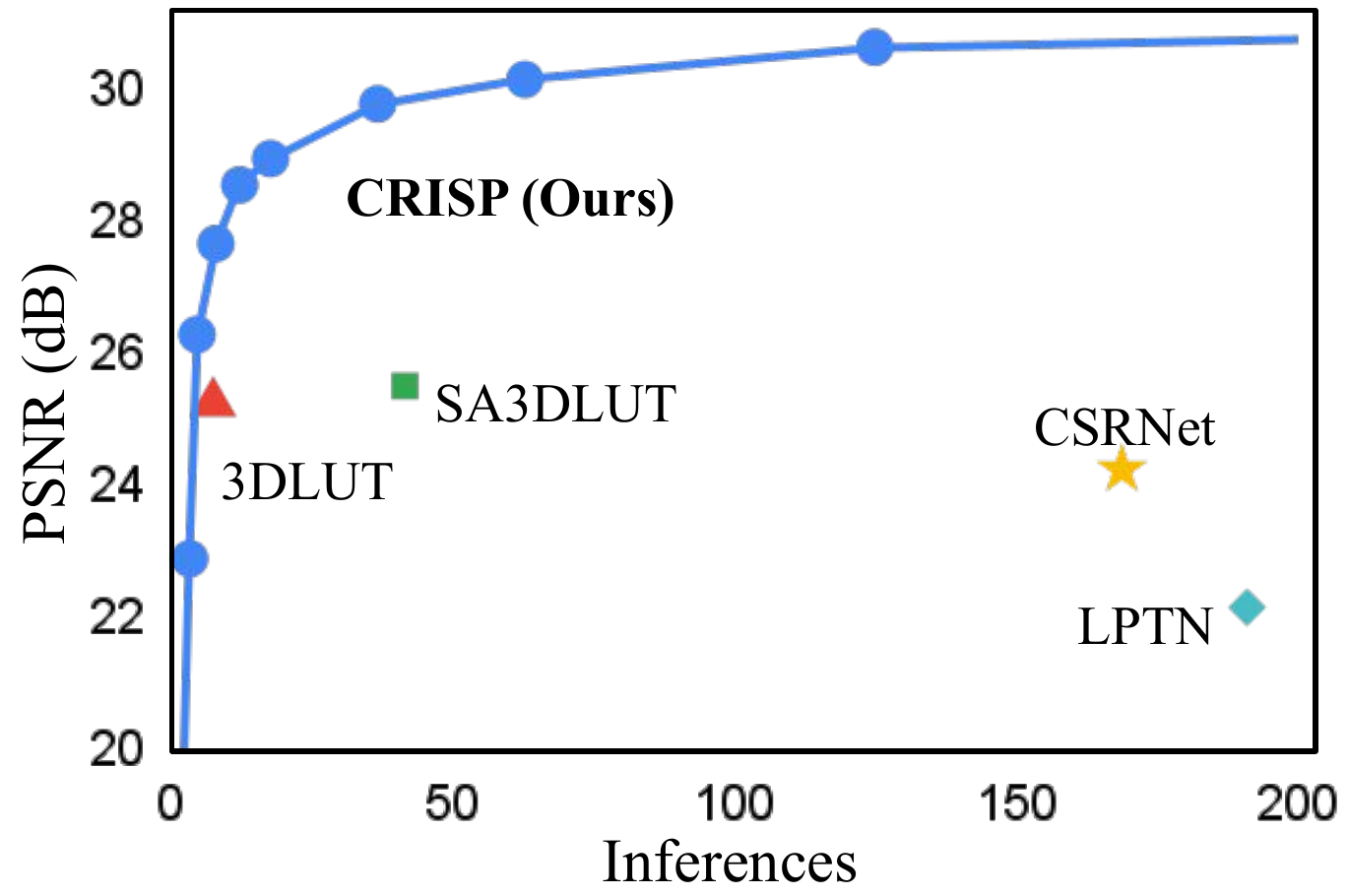}\\  
	\scriptsize
	\captionof{figure}{PSNR \textit{vs.} \#Inferences. CRISP can control the output image quality (PSNR) by increasing \#inferences. \#inferences for the other methods is the FLOPs multiplier to CRISP for visualization.
	}
	\vspace{-1em}
	\label{fig:greedysearch}
\end{minipage}
\end{table}
\begin{table}[t]
    	
\vspace{-0.2cm}
\end{table}


\subsection{Analysis}

\begin{figure*}[t]
	\centering\tiny
	\vspace{-0.3cm}
    \scalebox{1}{
    \hspace{-0.2cm}
	\setlength\tabcolsep{0pt}
    \begin{tabular}{cccccc} 
    ~~~~~~~~~~~~~~~~~~~~~~~~&~~~~~~~~~~~~~~~~~~~~~~~~&~~~~~~~~~~~~~~~~~~~~~~~~~~&~~~~~~~~~~~~~~~~~~~~~~~~~&~~~~~~~~~~~~~~~~~~~~~~~~~&~~~~~~~~~~~~~~~~~~~~~~~~~ \\
	 Input & (a) Fixed & (b) Input-adaptive  & (c) Task-adaptive & (d) Upper-bound & Expert
	\end{tabular}}\\	
	\includegraphics[width=1\linewidth]{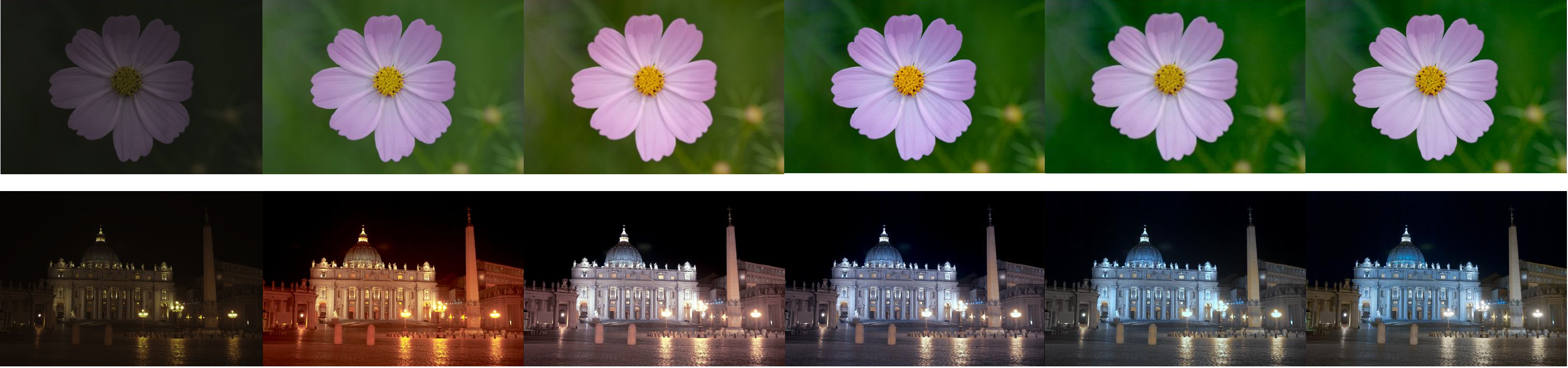}\vspace{-0.3cm}
	\scriptsize
	\caption{Qualitative comparison with different types of the 19-parameter ISP pipeline. 
	}
	\vspace{-1em}
	\label{fig:ablation_types}
\end{figure*}

\textbf{Effectiveness of task-adaptive parameters.}
CRISP uses \textit{task-adaptive} ISP parameters where tasks represent photo retouching skills and can be controlled by users.
To evaluate the effectiveness of the task-adaptation, we set the following types of models which share our ISP pipeline described in Section~\ref{sec:pipeline}:
%
\begin{itemize}\setlength\itemsep{-0.1em}
    \item \textbf{Fixed params.} Without the reparametrization, the ISP parameters $\boldsymbol{\phi}$ are directly optimized in training and have fixed values in testing.
    \item \textbf{Input-adaptive params.} We modify our encoder to use only low-quality images as input for both training and testing. The input low-quality images adapt the ISP parameters. 
    \item \textbf{Task-adaptive params.} CRISP is the model of task-adaptive params with the control dimension ($D$) of 3.
    \item \textbf{Upper-bound.} Using the architecture of \textit{fixed params}, we optimize the ISP parameters for each test image.
\end{itemize}
Figure~\ref{fig:ablation_types} and Table~\ref{table:ablation_tyeps} present the qualitative and quantitative results of the models.
The upper-bound achieves outstanding performances that indicate the great potential of a 19-parameter ISP pipeline for image enhancement.
By contrast, fixed params performs a 15 dB lower in PSNR than the upper-bound with the wrong white balance under the colored illumination (See Figure~\ref{fig:ablation_types}{\color{red}(a)}).
Input-adaptive params alleviate the effect of the colored illumination and some objects (\textit{e.g.}, the outer wall in Figure~\ref{fig:ablation_types}{\color{red}(b)}) look like ``true'' color.
However, the outputs have different styles from the images from Expert, resulting in a significant low PSNR.
The task-adaptive params produce results that are qualitatively (Figure~\ref{fig:ablation_types}{\color{red}(c)}) and quantitatively (Table~\ref{table:ablation_tyeps}) closer to upper-bound than others.

\textbf{Effectiveness of control dimension.}
Table~\ref{fig:ablation_control_dim} presents an ablation study to the control dimension ($D$).
While a single control dimension ($D$=$1$) improves only marginal PSNR than input-adaptive params (23.12 dB), the high flexibility to control ($D\geq 3$) performs accurate image retouching, where $D=3$ can be Pareto efficiency.
Interestingly, the model with a higher control dimension (\textit{e.g.}, $D=64$) than ISP parameters (19) still performs lower PSNR than upper-bound (31.47 dB \textit{vs.} 34.95 dB).
The performance gap comes from the generalization power of the decoder and the disentanglement power of the encoder.

\begin{table}[t]

\begin{minipage}[c]{0.5\textwidth}%
\centering\scriptsize
		\captionof{table}{Quantitative comparison with different types of the parametric ISP pipeline.}
    	\label{table:ablation_tyeps}
    	\vspace{-0.0cm}\hspace{-0.3cm}
    	\scalebox{1}{
    	\setlength\tabcolsep{4pt}
        	\begin{tabular}{lcc} 
                \toprule[1.5pt]
                Type  & PSNR & SSIM \\
        		\midrule[1.0pt]
                Fixed params & 19.98 & 0.8261 \\
                Input-adaptive params  & 22.72 & 0.8655 \\
                Task-adaptive params (ours) & 29.61 & 0.9199\\
        		\midrule[1.0pt]
                Upper-bound &  34.95 & 0.9500\\
                \bottomrule[1.5pt]
        	\end{tabular}} 
\end{minipage}
~~
\begin{minipage}[c]{0.45\textwidth}%
	\centering\scriptsize
	\vspace{0.4cm}
	\includegraphics[width=0.65\linewidth]{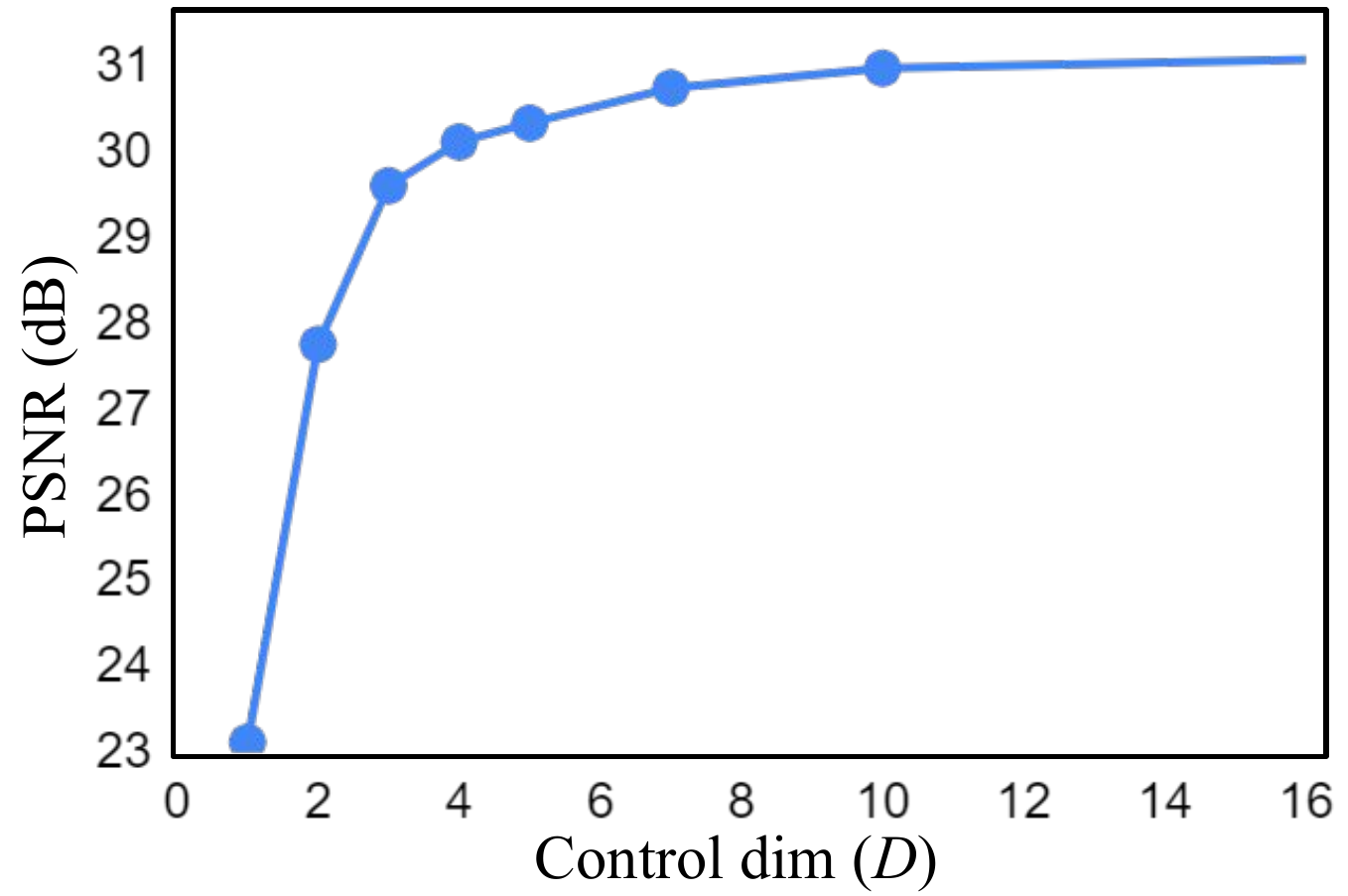}\\  
	\scriptsize
	\captionof{figure}{Ablation stody on control dimension ($D$) for task-adaptive params.
	}
	\vspace{-1em}
	\label{fig:ablation_control_dim}
\end{minipage}
\end{table}
\begin{table}[t]
    	
\vspace{-0.2cm}
\end{table}

\begin{figure}[h!]
\centering
    \footnotesize
        \vspace{-0.1cm}
    \includegraphics[width=1\linewidth]{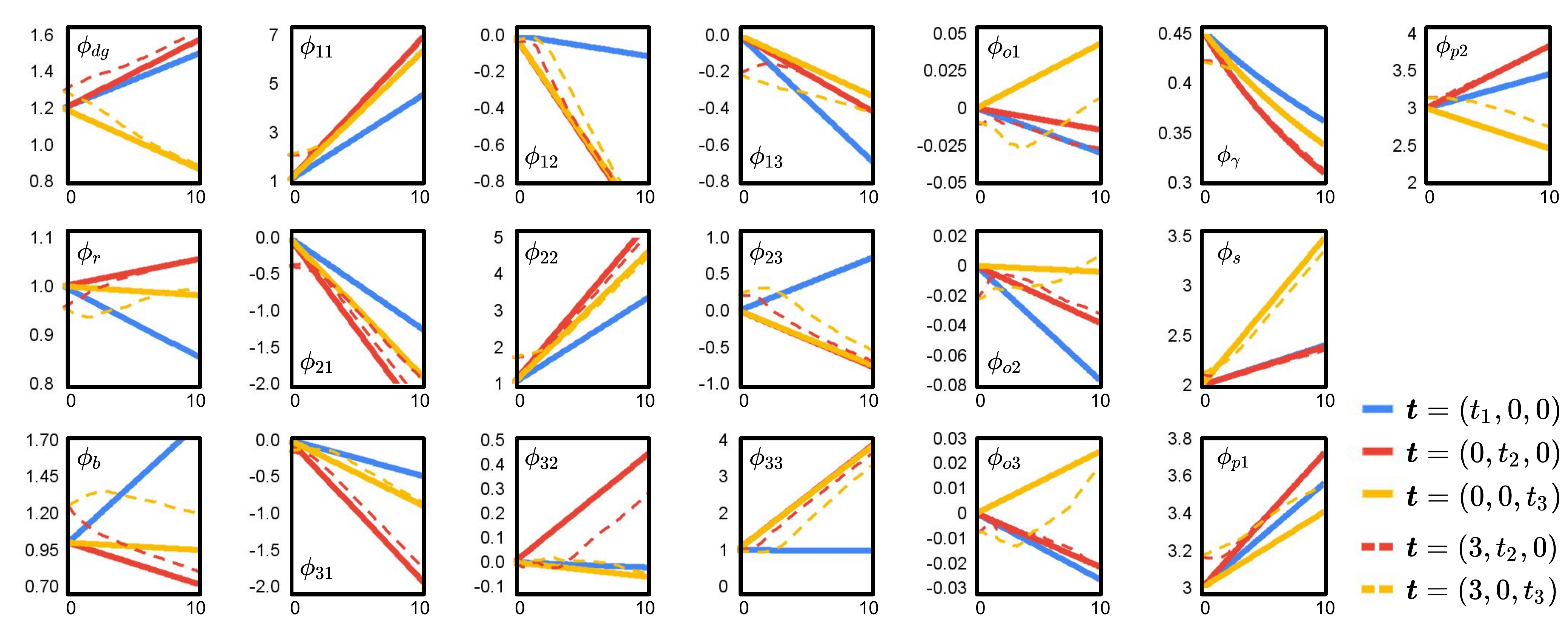}\\
    \vspace{-0.3cm}
    \caption{\footnotesize Graphs of the 19 ISP parameters to task vector. $t_d$ is the variable of $x$-axis.  
    }
    \vspace{-0.0cm}
    \label{fig:isp_params}
\end{figure}

\begin{figure}[h!]
\centering
    \scriptsize
    \setlength\tabcolsep{0.5pt}
    \color{white}
	\begin{tabular}{lllllllll}
    \includegraphics[height=0.1\linewidth]{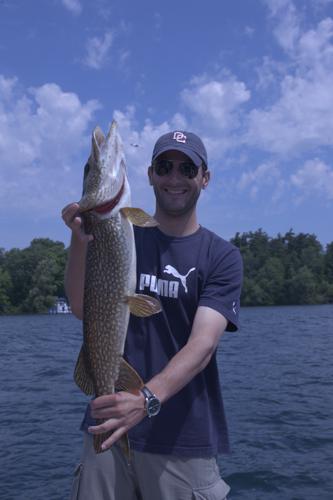} & \includegraphics[height=0.1\linewidth]{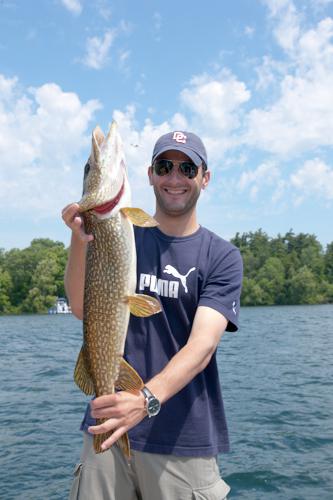} &
    \includegraphics[height=0.1\linewidth]{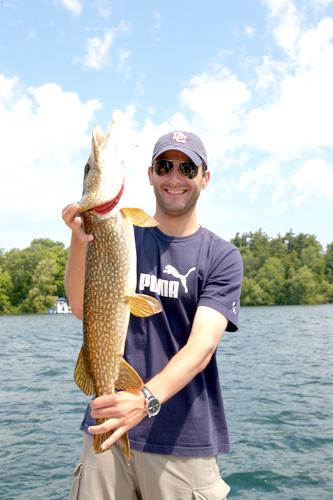} &
    \includegraphics[height=0.1\linewidth]{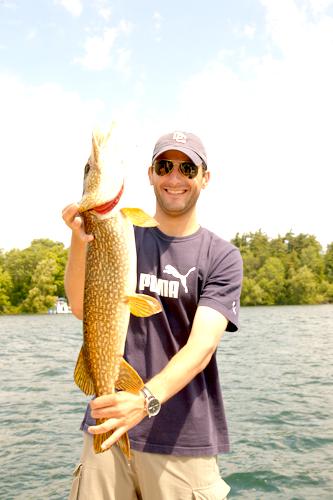} & ~~~~ &
    \includegraphics[height=0.1\linewidth]{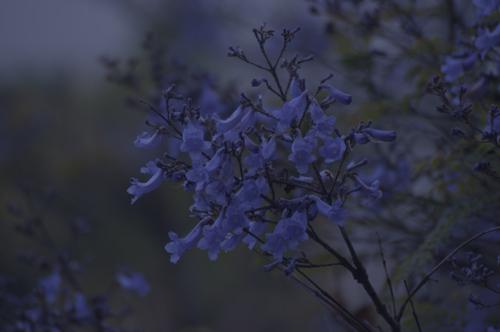} &
    \includegraphics[height=0.1\linewidth]{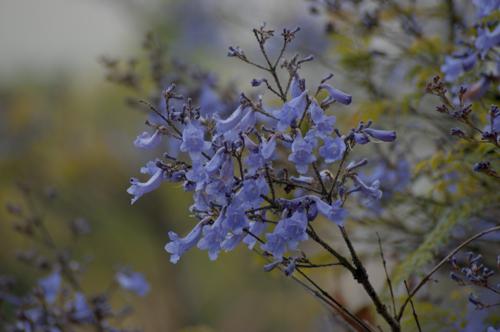} &
    \includegraphics[height=0.1\linewidth]{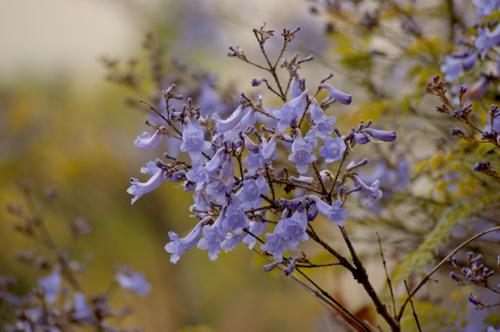} &
    \includegraphics[height=0.1\linewidth]{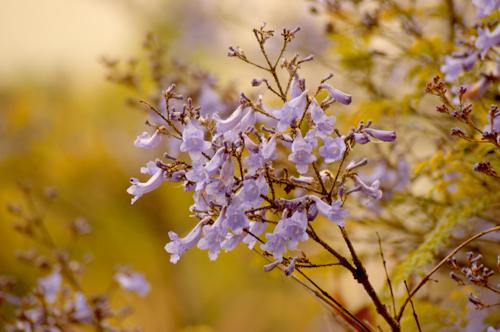}     \vspace{-0.35cm} \\
    (3,0,6)& (3,3,6) &(3,6,6) & (3,9,6)  & & (3,0,6)& (3,3,6) &(3,6,6) & (3,9,6) \\
    \includegraphics[height=0.1\linewidth]{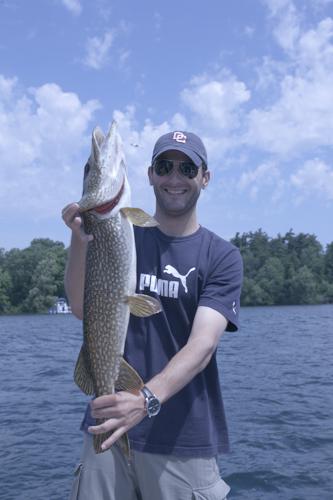} & \includegraphics[height=0.1\linewidth]{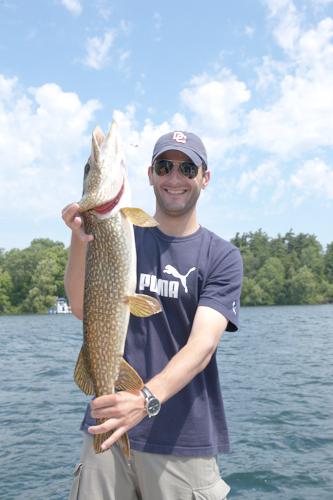} &
    \includegraphics[height=0.1\linewidth]{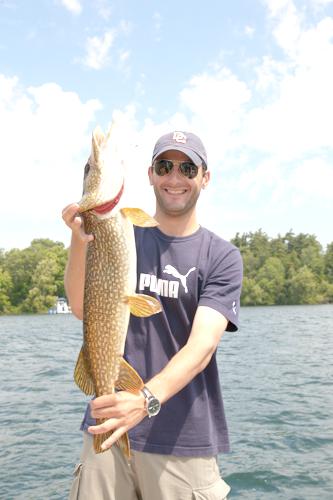} &
    \includegraphics[height=0.1\linewidth]{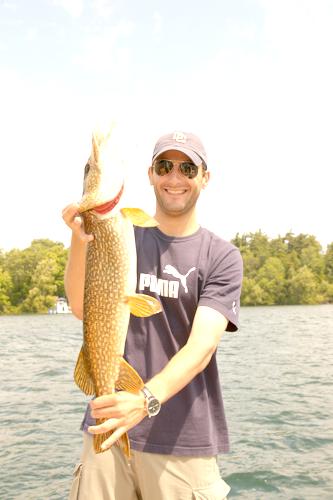} & ~~~~ &
    \includegraphics[height=0.1\linewidth]{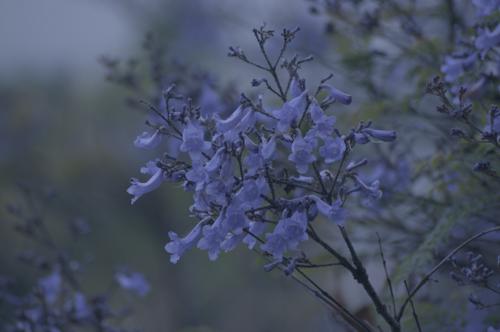} &
    \includegraphics[height=0.1\linewidth]{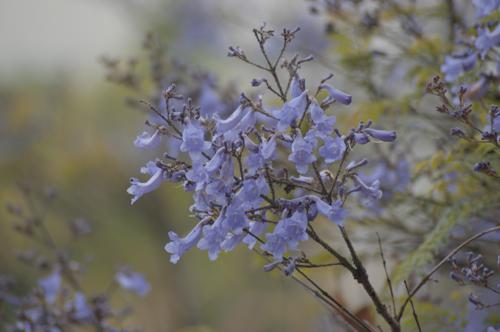} &
    \includegraphics[height=0.1\linewidth]{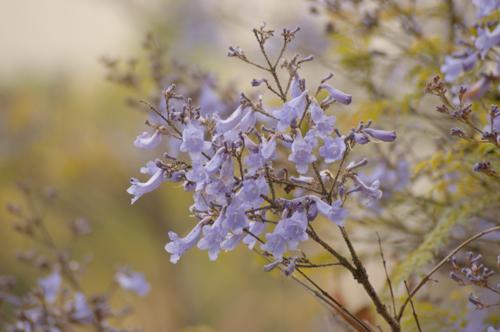} &
    \includegraphics[height=0.1\linewidth]{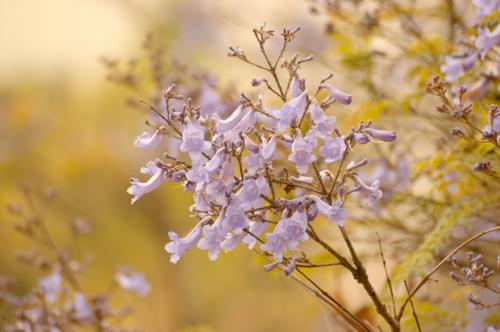} \vspace{-0.35cm} \\
    (3,0,3)& (3,3,3) &(3,6,3) & (3,9,3) & & (3,0,3)& (3,3,3) &(3,6,3) & (3,9,3) \\
    \includegraphics[height=0.1\linewidth]{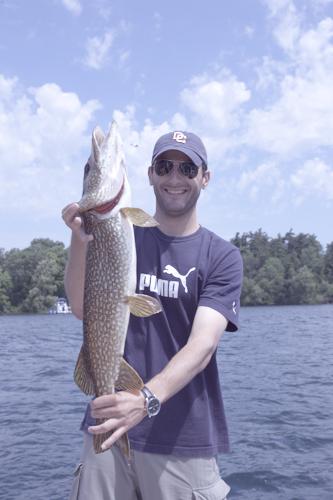} & \includegraphics[height=0.1\linewidth]{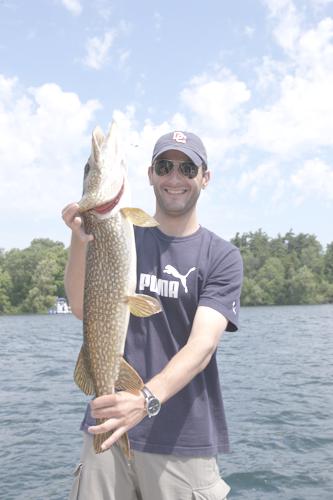} &
    \includegraphics[height=0.1\linewidth]{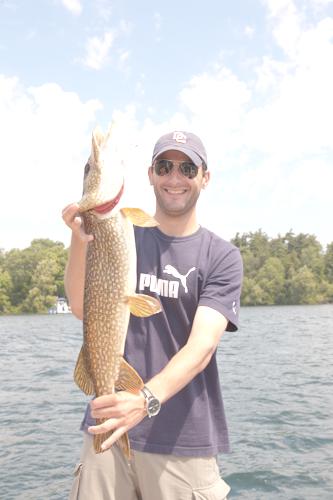} &
    \includegraphics[height=0.1\linewidth]{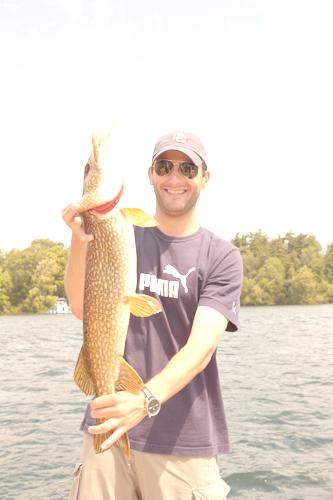} & ~~~~ &
    \includegraphics[height=0.1\linewidth]{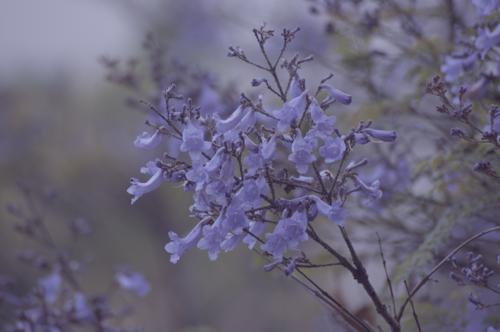} &
    \includegraphics[height=0.1\linewidth]{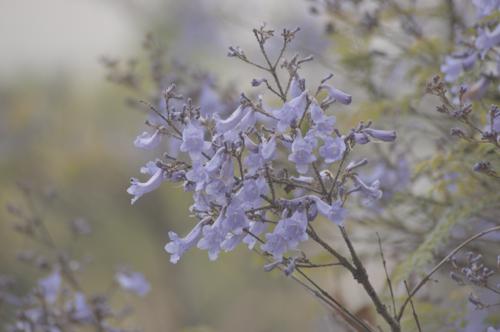} &
    \includegraphics[height=0.1\linewidth]{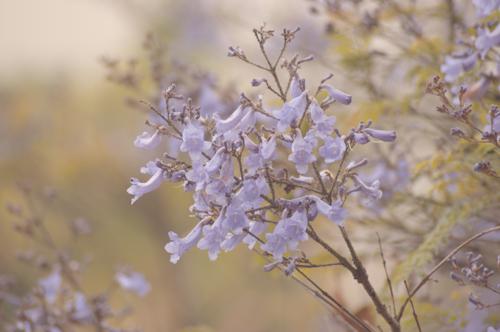} &
    \includegraphics[height=0.1\linewidth]{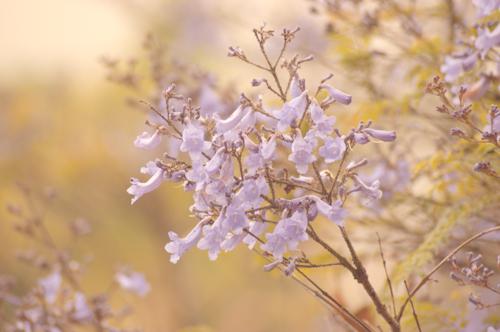} \vspace{-0.35cm} \\
    (3,0,0)& (3,3,0) &(3,6,0) & (3,9,0) & & (3,0,0)& (3,3,0) &(3,6,0) & (3,9,0) \\
    \end{tabular}
    \vspace{-0.3cm}
    \caption{\footnotesize Consistent style generation. $(\cdot,\cdot,\cdot)$ denotes the task vector $\boldsymbol{t}$. CRISP generates consistent retouching styles over input images for the same task vectors.}
    \label{fig:consistency}
\end{figure}

\textbf{Controllability of CRISP.}
As Figure~\ref{fig:greedysearch} demonstrates that a simple greedy search algorithm can find high quality images in few steps, CRISP is easy to control.
To analyze its controllability further, Figure~\ref{fig:isp_params} visualizes values of ISP parameters while adjusting a dimensional value of task vectors.
Solid lines are generally linear functions since the other values in task vectors are zero.
Dashed lines draw non-linear functions from the fully connected layers ($g$).
For both cases, task vectors have an \textit{insensitive and distinct tendency} to change ISP parameters. 
Moreover, CRISP generates a retouching style \textit{independently} to the input image.
Figure~\ref{fig:consistency} illustrates consistent style generation over two input images for the same task vectors.
Although the two images render different colors for a task vector, the retouching tendency is maintained while adjusting the values of the task vector.
Therefore, users can easily catch retouching styles of task vectors by adjusting a few examples.

\section{Conclusion}
\label{sec:conclusion}
We present a semi-automatic algorithm, called CRISP, for image enhancement.
While existing automated algorithms aim to generate an universal high-quality (HQ) photograph for a test image, CRISP can generate multiple styles of HQ images.
CRISP consists of only 19 parameters for image processing, bringing the enhancement process computationally efficient.
Experiments demonstrate that CRISP achieves better image quality than SotA methods in MOS, PSNR, and SSIM while generating multiple styles of HQ images.
We analyze the effectiveness and efficiency of controllability compared to the universal automatic approach.

%
%
\bibliographystyle{splncs04}
\bibliography{egbib}



\section*{Supplementary material}

# Controllable Image Enhancement
## – *Supplementary Document* –

Heewon Kim and Kyoung Mu Lee

ASRI, Department of ECE, Seoul National University
{ghimhw, kyoungmu}@snu.ac.kr

This document presents the network architecture of CRISP in Section S1, additional experiments in Section S2, and additional qualitative results used for user study in Section S3.

## S1 Neural Network Architecture

Figure S1 presents the encoder-decoder architecture of CRISP described in Figure 3. The encoder uses convolutional layers to disentangle retouching skills during training. The decoder consists of fully connected layers with the feature resolution of 1×1 that are computationally efficient.

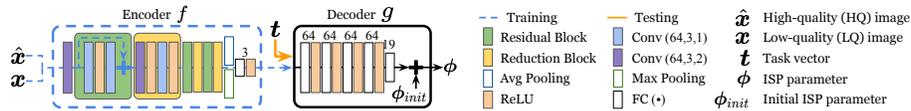

Fig. S1: Architectures of our encoder-decoder framework. Conv (·,·,·) denotes convolution layer with the number of filters, the kernel size, and the value of stride, respectively. Avg pooling and Max pooling reduce the feature resolution to 1×1. FC (·) is a fully connected layer, where · is the number of filters denoted at the top of the boxes.

## S2 Additional Experiment

The proposed training scheme allows a model to use multiple high-quality (HQ) images for a single low-quality (LQ) image. Table S1 demonstrates the performances of CRISP trained with 5 HQ images from 5 experts for a LQ image.

Table S1: Effectiveness of training with multiple styles of HQ images on MIT-Adobe FiveK. Reported scores are PSNR (dB). Expert (All) denotes 5 styles from 5 experts.

|            | CSRNet   | CRISP    |              |
|------------|----------|----------|--------------|
| Train data | Expert-C | Expert-C | Expert (All) |
| Expert-A   | 24.15    | 28.29    | **29.98**    |
| Expert-B   | 24.95    | 30.60    | **32.06**    |
| Expert-C   | 24.98    | **29.61**| 29.11        |
| Expert-D   | 22.73    | 29.28    | **30.73**    |
| Expert-E   | 21.74    | **29.69**| 29.65        |

## S3 Additional Qualitative Results

Figure S2 and S3 visualize examples used for user study of single-style MOS. Users scored the quality of each image result. Figure S4, S5, S6, and S5 demonstrate images used for user study of multiple-style MOS. Users selected the most preferred image in 5 candidates and scored the quality of the selected images.

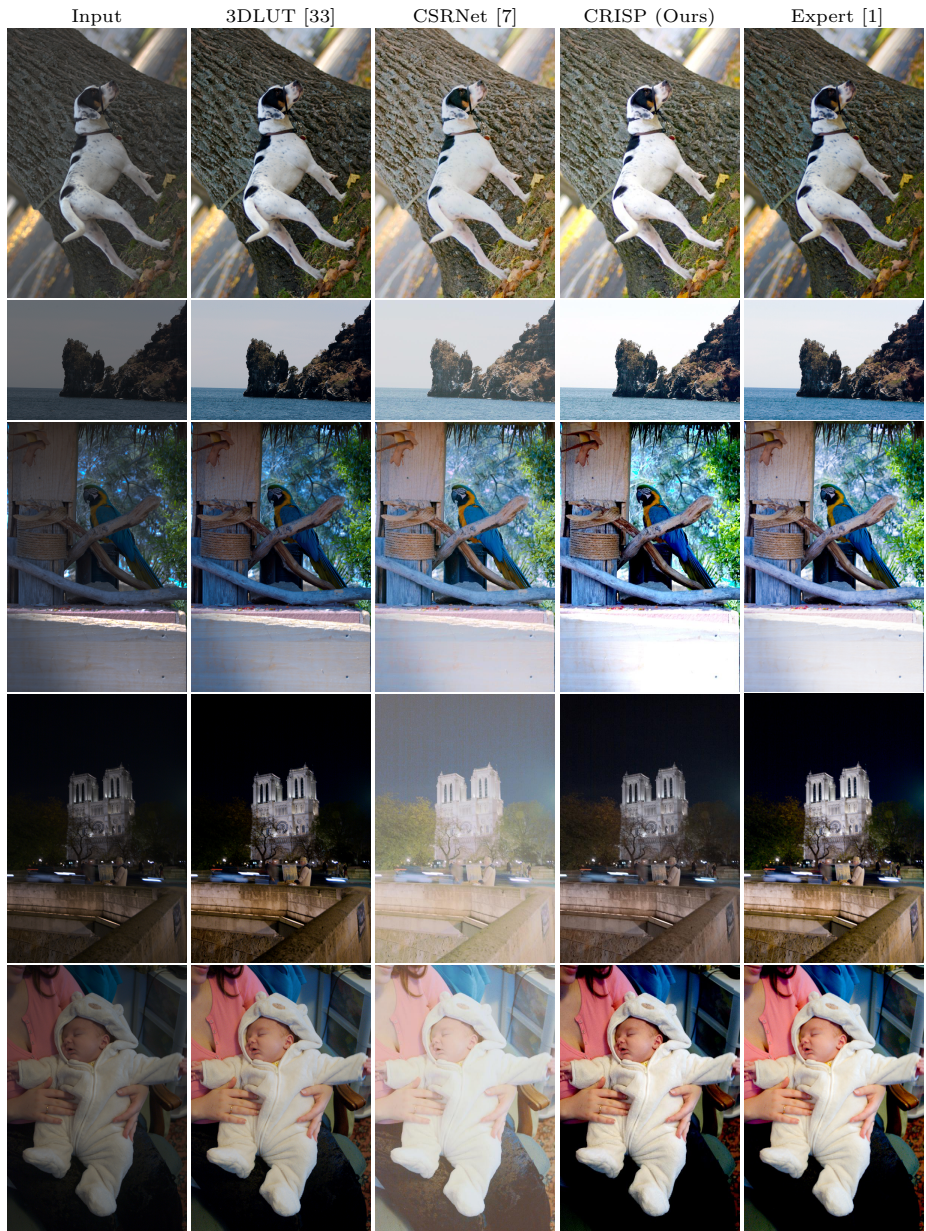

Fig. S2: Example images for user study of single-style MOS.

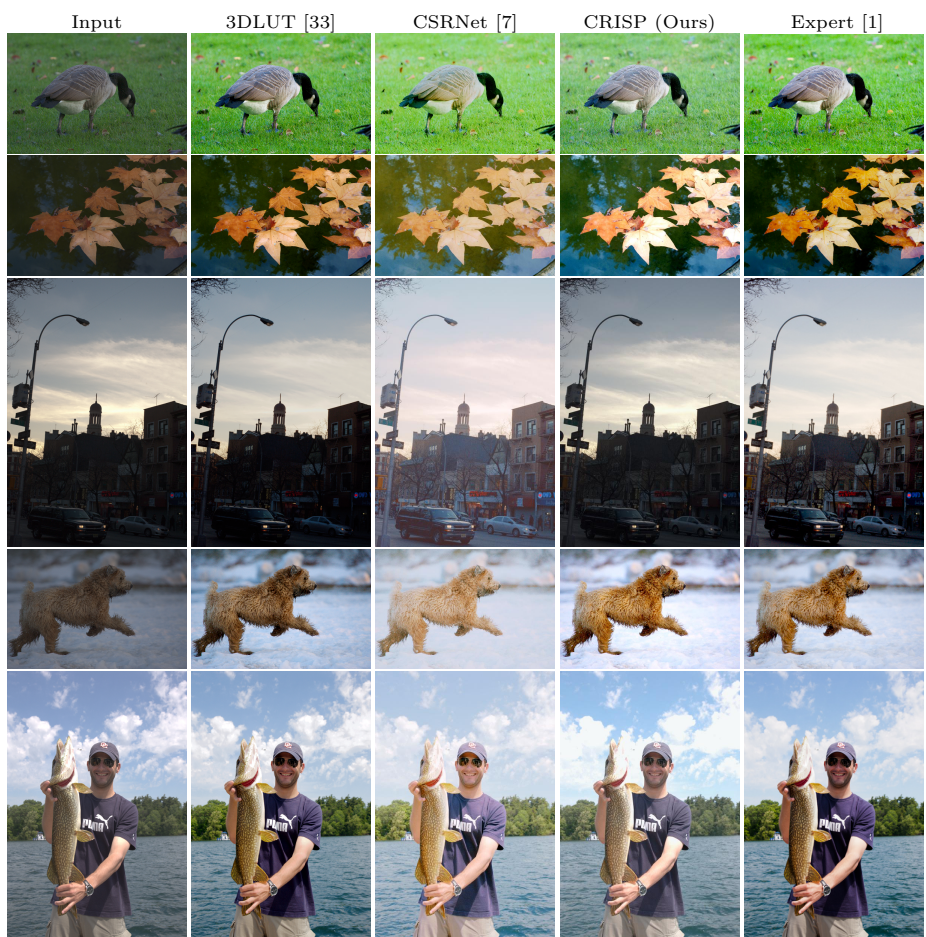

Fig. S3: Example images for user study of single-style MOS.

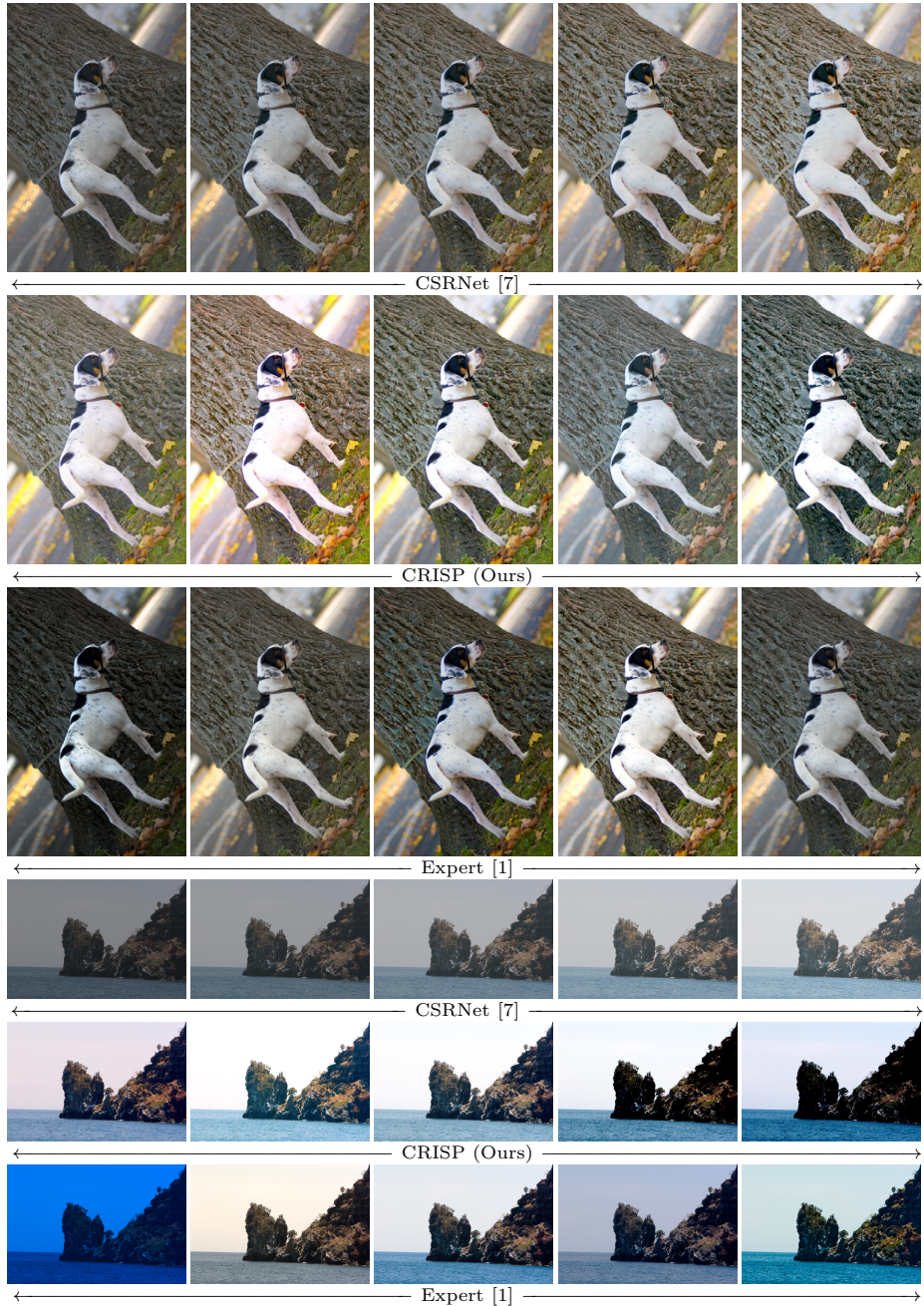

Fig. S4: Example images for user study of multiple-style MOS.

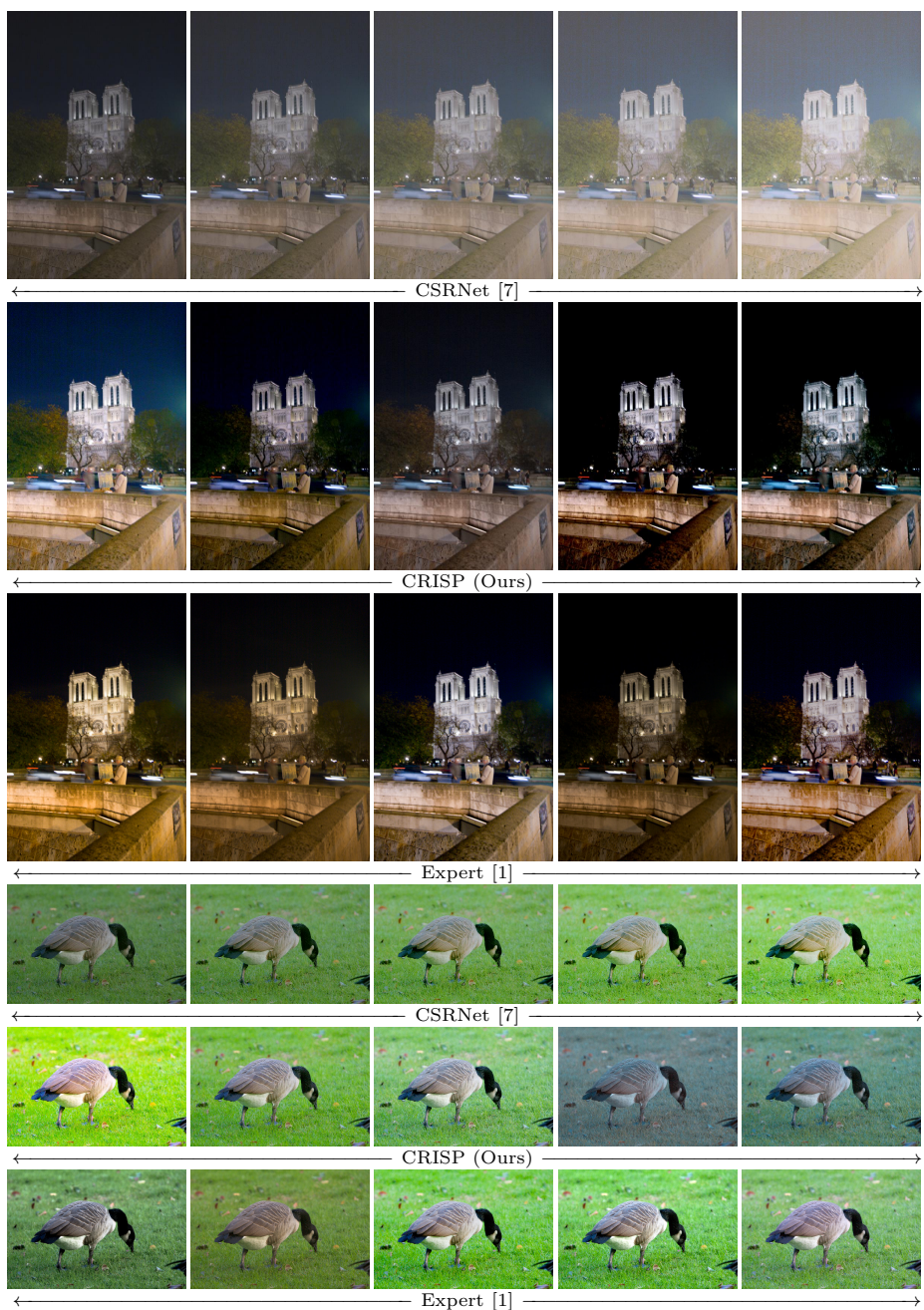

Fig. S5: Example images for user study of multiple-style MOS.

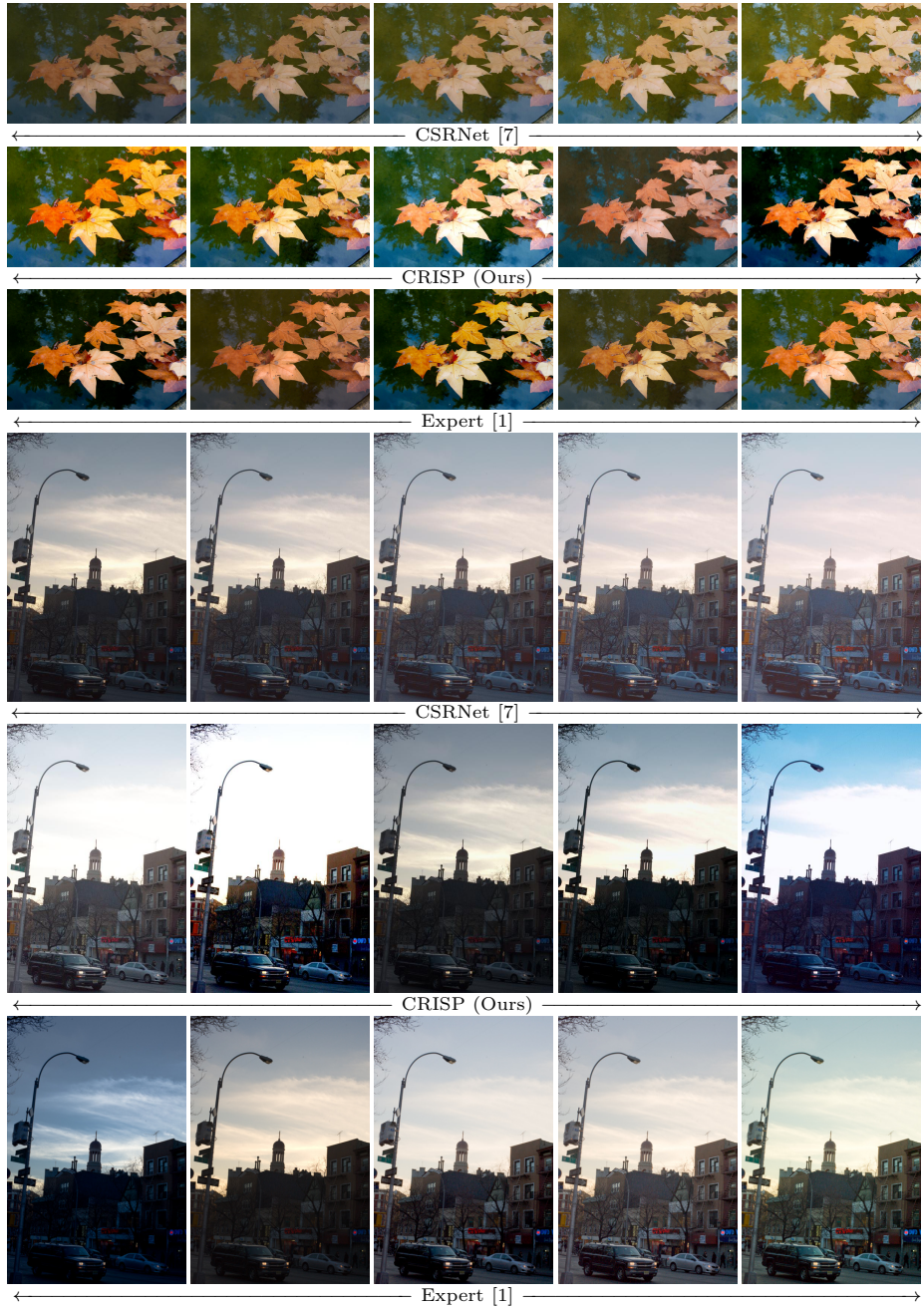

Fig. S6: Example images for user study of multiple-style MOS.

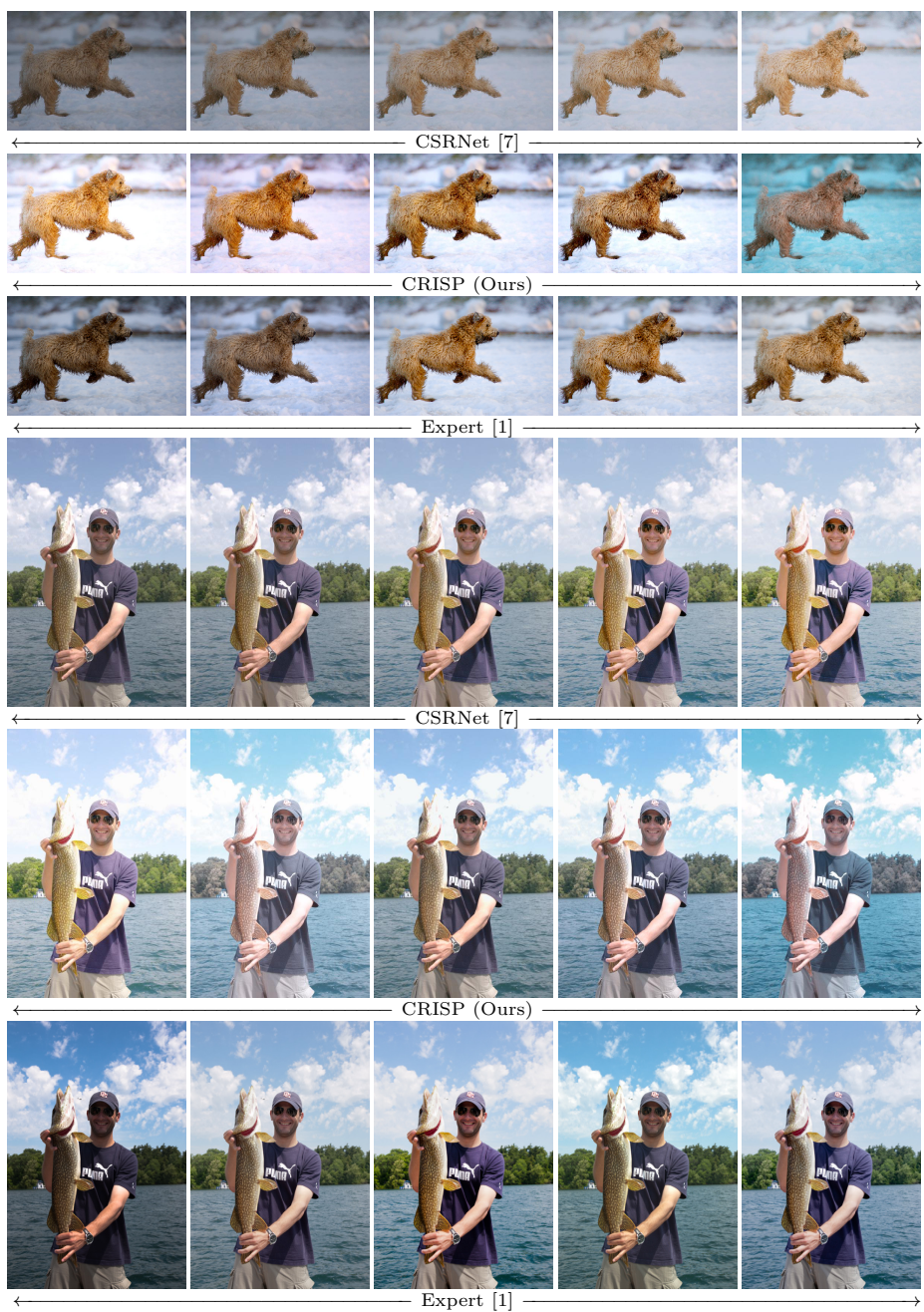

Fig. S7: Example images for user study of multiple-style MOS.



\end{document}